\documentclass[journal]{IEEEtran}
\usepackage{graphicx}
\usepackage{amsmath}
\usepackage{makecell}
\usepackage{multirow}
\usepackage{collcell}
\usepackage{colortbl}
\usepackage{bm}
\usepackage{tikz}
\usepackage{pgfplots}
\usepackage{pgfplotstable}
\usepackage{enumitem}
\usepackage{adjustbox}
\usepackage{soul}
\usepackage{newfloat}
\usepackage{xcolor}
\usepackage{siunitx}
\usepackage{collcell}
\usepackage{graphicx} 
\usepackage{tikz}
\usepackage{subcaption}
\usepackage{pgfplots}
\usepackage{mathtools}
\usepackage{multirow}
\usepackage{url}
\usepackage[font=small]{caption}

\usepackage[T1]{fontenc}  
\usepackage{amsmath,amsfonts,amssymb,amsthm}
\DeclareFloatingEnvironment[fileext=dia, within=section]{diagram}
\usepackage[normalem]{ulem}
\useunder{\uline}{\ul}{}
\usepackage{tkz-kiviat}
\usetikzlibrary{arrows}
\usetikzlibrary{patterns}
\DeclareRobustCommand{\legendsquare}[1]{%
  \tikz[baseline=(a.south)]{\node[#1, inner sep=.8ex, outer sep=0] (a) {};}%
}

\newcommand{\TrainingSet}{\bm{T}}

\newcommand{\TestSet}{\Psi}

\newcommand{\AllBands}{\lambda}

\newcommand{\NumberOfClasses}{c}

\newcommand{\PatchSize}{p}
\usepackage{cleveref}

\usetikzlibrary{arrows,positioning}
\usetikzlibrary{plotmarks}
\usetikzlibrary{calc}
\usepgfplotslibrary{groupplots}
\usetikzlibrary{patterns}
\usetikzlibrary{shapes.geometric}
\pgfdeclarelayer{bg}    % declare background layer
\pgfsetlayers{bg,main, background}
\usepgfplotslibrary{fillbetween}
\pgfplotsset{compat=1.17, compat/show suggested version=false}
\def\Decimal{0.0000}% This structure should corresponds to the ".2" of "table-format"

\def\Ulinehelp#1.#2 {%
  #1.#2\setbox0=\hbox{#1\Decimal}\hspace{-\wd0}{\if\relax#2\relax%
    \uline{\phantom{#1.0}}\else\uline{\phantom{#1.#2}}\fi}%
}
\usetikzlibrary{patterns}
\DeclareRobustCommand{\legendsquare}[1]{%
  \tikz[baseline=(a.south)]{\node[#1, inner sep=.8ex, outer sep=0] (a) {};}%
}
\usetikzlibrary{spy}
\usepackage{algorithm}
\usepackage{algpseudocode}

\ifCLASSINFOpdf
\else
\fi

\definecolor{hotmagenta}{rgb}{1.0, 0.11, 0.81}
\definecolor{gray1}{RGB}{0,0,0}
\definecolor{gray2}{RGB}{80,80,80}
\definecolor{gray3}{RGB}{140,140,140}
\definecolor{gray4}{RGB}{170,170,170}
\definecolor{gray5}{RGB}{200,200,200}
\definecolor{gray6}{RGB}{220,220,220}
\definecolor{coordscolor}{RGB}{235,235,235}

\begin{document}
%\title{Hyperspectral Data Augmentation}
%\title{Training/Test-Time Hyperspectral Data Augmentation}
% \title{Hyperspectral Unmixing Using Multi-Branch Convolutional Neural Networks}
% \title{Validating Multi-Branch Convolutional Neural Networks for Hyperspectral Unmixing}
% \title{Multi-Branch Deep Network and its Unbiased Validation for Hyperspectral Unmixing}
% \title{A Multi-Branch Network for Hyperspectral Unmixing and }
% \title{A new data fusion scheme for hyperspectral unmixing with unbiased validation}
% \title{A new data fusion scheme for hyperspectral unmixing with leak-free validation}
\title{A Multibranch Convolutional Neural Network for Hyperspectral Unmixing}
\author{
Lukasz Tulczyjew, Michal Kawulok,~\IEEEmembership{Member,~IEEE}, Nicolas Long\'{e}p\'{e}, Bertrand Le Saux,~\IEEEmembership{Senior Member,~IEEE}, and Jakub Nalepa,~\IEEEmembership{Member,~IEEE}
\thanks{LT, MK and JN are with Silesian University of Technology (SUT), Gliwice, Poland (e-mail: jnalepa@ieee.org) and with KP Labs, Gliwice, Poland. NL and BLS are with $\Phi$-lab, European Space Agency, Frascati, Italy.\\This work was funded by the European Space Agency (the GENESIS project), supported by the ESA $\Phi$-lab ({https://philab.phi.esa.int/}), and by the SUT grant for maintaining and developing research potential.}% <-this % stops a space
}

\markboth{IEEE GEOSCIENCE AND REMOTE SENSING LETTERS}%
{Shell \MakeLowercase{\textit{et al.}}: Bare Demo of IEEEtran.cls for IEEE Journals} 
 
\maketitle

\begin{abstract}

Hyperspectral unmixing remains one of the most challenging tasks in the analysis of such data. Deep learning has been blooming in the field and proved to outperform other classic unmixing techniques, and can be effectively deployed onboard Earth observation satellites equipped with hyperspectral imagers. In this letter, we follow this research pathway and propose a multi-branch convolutional neural network that benefits from fusing spectral, spatial, and spectral-spatial features in the unmixing process. The results of our experiments, backed up with the ablation study, revealed that our techniques outperform others from the literature and lead to higher-quality fractional abundance estimation. Also, we investigated the influence of reducing the training sets on the capabilities of all algorithms and their robustness against noise, as capturing large and representative ground-truth sets is time-consuming and costly in practice, especially in emerging Earth observation scenarios.

\end{abstract}


\IEEEpeerreviewmaketitle

\section{Introduction} \label{sec:intro}

%\subsection{Related Work}

%\subsubsection{Hyperspectral Image Segmentation}\label{sec:segmentation}

Hyperspectral imaging (HSI) allows for retrieving data of high spectral dimensionality, commonly at a cost of lower spatial resolution, which means that a single hyperspectral pixel presents a mixture of  signatures from many endmembers, or pure signature of a given material. This is especially visible for the satellite missions, where the spatial resolution may reach tens of meters for such spaceborne-acquired hyperspectral data. The process of estimating the individual endmembers along with their fractional abundances is known as \emph{hyperspectral unmixing} (HU). Initial approaches toward HU, including the linear mixing model (LMM)~\cite{heinz1999fully}, were underpinned with the assumption that each pixel is a linear combination of the endmembers' abundances. Although this assumption may not hold due to the variations in illumination, atmospheric conditions, and spectral variability, resulting in low accuracy of the linear models, the linear
mixture model is a good approximation for remote sensing applications, in which hyperspectral pixels commonly contain large and homogeneous regions of coherent materials~\cite{Rasti2021undip}. There are machine learning models, including support vector regression (SVR)~\cite{6638039}, as well as artificial neural networks~\cite{5967899}, which rely on the training data for HU, but they can also be enriched with certain priors based on physical modeling~\cite{Xiong2021}. We have been observing an unprecedented success of deep learning in various fields of science and industry, with HU not being an exception here. The process of estimating the fractional abundances can be effectively learned from the data using convolutional neural networks (CNNs)~\cite{cnn_unmixing} which benefit from automated representation learning and can capture features that would be difficult or impossible to extract using hand-crafted feature extractors. Also, the deep image prior was recently exploited for the unmixing task~\cite{Rasti2021undip}---here, the authors additionally addressed the problem of the endmember estimation, which is another important challenge in HU, especially relevant to large-scale Earth observation scenarios.

\begin{figure}[t!]
    \centering
    \includegraphics[width=\columnwidth]{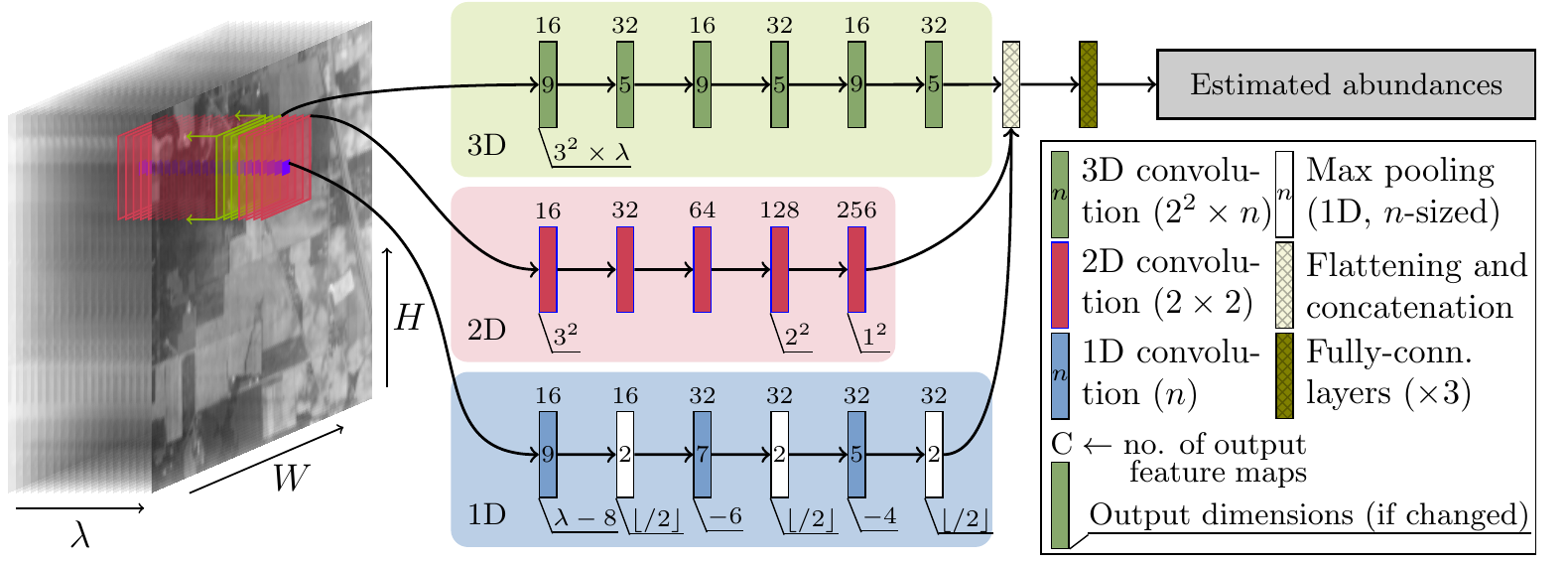}
    \caption{An outline of our multi-branch architecture. Three branches extract spectral (1D), spatial (2D) and spectral-spatial (3D) features from an input HSI, and they are fused to estimate the abundances. For the 1D branch, we show a relative change in the dimensionality.}
    \label{fig:net}
\end{figure}
%For our implementation, see \url{https://gitlab.com/jnalepa/mbhu}.

There are two important research directions in the intensively explored field of HU using deep learning~\cite{9324546}: (\textit{i})~developing new deep architectures and (\textit{ii})~attempts to overcome the problem of limited ground-truth data. The latter can be addressed with semi-~\cite{8898846}, weakly-supervised~\cite{weaklyae}, and unsupervised learning~\cite{RastiKoirala2021}. In~\cite{8984691}, a deep convolutional autoencoder (DCAE) was proposed for a supervised unmixing scenario, but DCAEs are now commonly exploited in unsupervised approaches~\cite{Shahid2021}. Recently, Jin et al. trained an unsupervised DCAE in an adversarial manner to increase its robustness against noise~\cite{JinMa2021}. In~\cite{weaklyae}, a weakly-supervised autoencoder (WS-AE) was introduced which requires a small portion of the labeled set. These architectures extract either spectral or spectral-spatial features from an input HSI cube, with the latter being reported to obtain better performance~\cite{8984691}. 

However, it was not attempted to combine the spatial and spectral features later in the processing chain. Such late fusion approaches---implemented in a form of multi-branch networks---were developed for other tasks related with HSI analysis~\cite{8738045}, but they were not reported for HU so far. They were utilized in hyperspectral classification~\cite{GAO201911}, and such techniques encompass, among others, attention multi-branch CNNs benefiting from the adaptive region search~\cite{9160985}, multi-branch-multi-scale residual fusion networks~\cite{10.1117/1.JRS.15.024512}, and multi-branch networks based on weight sharing~\cite{9553652}.

We address this research gap by introducing a new multi-branch convolutional architecture for the hyperspectral unmixing which benefits from the early and late fusion of spectral and spatial features to deliver accurate fractional abundances (Section~\ref{sec:method}). In a high-level flowchart in Fig.~\ref{fig:net}, we highlight the operation of the proposed deep learning model trained in a supervised way, utilizing the ground-truth abundances (hence, we do \textit{not} tackle another important problem of the endmember determination~\cite{Rasti2021undip}). The 1D and 2D branches extract spectral and spatial features, respectively, and the 3D branch realizes an early fusion of the features extracted from these two domains. Afterwards, the features extracted with these branches are combined within the late fusion. In contrast to the networks based on the spatial-spectral features (early fusion), we ensure that the valuable features will be extracted from both spatial and spectral domains. Our thorough experiments (Section~\ref{sec:experiments}) indicate that the suggested CNN outperforms other state-of-the-art techniques over several widely-used benchmarks and leads to obtaining high-quality unmixing (consistently outperforming other techniques in both root mean square error and the root mean square abundance angle distance). We performed the ablation analysis to understand (\textit{i})~the impact of introducing the spectral, spatial, and spectral-spatial branches into the model, (\textit{ii})~the sensitivity of our techniques to noise, and (\textit{iii})~their performance for various sizes of the training sets. Finally, we make our implementation, together with the detailed CNN architectural diagrams available at \url{https://gitlab.com/jnalepa/mbhu} to ensure full reproducibility.

\section{Multi-Branch CNNs for HU}\label{sec:method}

% In this section, we discuss our multi-branch CNN for HU, together with its various architectural modifications in Section~\ref{sec:multibranch_for_hu}, whereas the proposed approach for elaborating training-test splits is presented in Section~\ref{sec:dataset_splits}.

% \subsection{Multi-Branch CNNs for HU}\label{sec:multibranch_for_hu}

In our \textbf{M}ulti-\textbf{B}ranch CNN (\textbf{MB}), we exploit spectral, spatial, and spectral-spatial features to improve the quality of HU (Fig.~\ref{fig:net}). The architecture embodies parallel feature extraction branches, where each branch can be portrayed as a separate module encapsulating several convolutional layers---the input constitutes a 3D hyperspectral patch of size $p \times p \times \AllBands$.

\subsection{Multi-Branch Feature Extraction}

The first block applies three 1D convolutional operations as well as three max pooling layers. The spectral extent of the filters is equal to 9, 7, and 5 for each layer, respectively, whilst the max pooling window encompasses two activation units. Since the input sample consists of three axes, the spatial dimensions are concatenated. Consequently, the input into the 1D convolutional layers is two-dimensional, and the number of pixels is treated as the number of input channels. Therefore, the input tensor is reshaped as: $\PatchSize\times \PatchSize\times\AllBands\rightarrow \PatchSize^2\times\AllBands$, where $\PatchSize\times\PatchSize$ is the input patch's size. As a result, the convolving kernel extracts the spectral representation of features by sliding along the spectral (band) dimension. In the 2D convolutional branch, we capture the \textit{spatial} features with five convolutional kernels of the $2\times 2\times \AllBands$ size, spanning the entire spectral dimension of the input patch of size $\PatchSize \times \PatchSize$. In the last 3D convolutional branch, we use three blocks, each one consisting of two layers with the $2\times 2\times9$ and $2\times 2\times 5$ kernels, to extract \textit{spectral-spatial} features. The output of every branch is concatenated and serves as the input to the fully-connected regression part that estimates the abundance fractions for each endmember. It incorporates 512, 64, and $\NumberOfClasses$ units in the consecutive layers, where $\NumberOfClasses$ denotes the number of endmembers in a scene. We utilize the Rectified Linear Unit (ReLU) activations.

\subsection{Extensions of the Architecture}

The baseline multi-branch CNN that was discussed in the previous section has been extended with the following modifications which will be experimentally analyzed in Section~\ref{sec:experiments}:
\begin{enumerate}[wide,leftmargin=*, labelindent=0pt, labelwidth=!]
    \item \textbf{Dimensionality reduction (MB-DR)}. As the number of features extracted in the 3D branch can be of orders of magnitude larger than those extracted in other branches (e.g.,~in MB, we would have ca. 500, 200, and more than $4.6\cdot 10^4$ features extracted in the 1D, 2D, and 3D branches for an input patch of size $3\times 3\times 162$), we introduce dimensionality reduction in the 3D branch. We exploit two additional 3D convolutional layers without padding, with the same kernel dimensionalities and larger strides, followed by a flattening layer before the concatenation to reduce the number of features resulting from the 3D branch, and to keep it comparable to other branches.
    \item \textbf{Residual connections}. Residual connections can significantly accelerate the process of training of deeper neural networks through improving the gradient flow and mitigating the vanishing and exploding gradient problem~\cite{resnets}. In the \textbf{MB-Res} model, we include the residual connections in the 3D branch which help us propagate the original hyperspectral information within the network. We include the dimensionality reduction mechanism from MB-DR---the skip connections bypass the extraction part of this branch and the original data characteristics are fused with the spectral-spatial features.
    \item \textbf{Sequential training strategy}. Here, MB-Res is trained in two steps. First, we train each branch separately to decouple them from other ones. Afterwards, we can either fine tune the entire architecture once it is combined into a multi-branch network, or train the regression part only. In the former case, the strategy may be understood as pre-training the feature extractor (MB-PT), whereas the latter strategy corresponds to the transfer learning-like scheme (MB-TL). Therefore, the main difference between MB-PT and MB-TL is the ``depth'' of the parameters' update---in MB-PT, we fine tune all trainable parameters, whereas in MB-TL, only a small subset of them.
\end{enumerate}

Although the approaches introduced in this section are model-agnostic and can be incorporated into other multi-branch CNNs, we analyze the models in the following order, with the networks expanding their predecessors: MB$\rightarrow$MB-DR$\rightarrow$MB-Res$\rightarrow$MB-PT and MB-TL. It will help track the impact of specific components on the pipeline's performance. %To ensure full reproducibility, the implementation of all introduced models, together with their detailed parameterization, is available at \url{https://gitlab.com/jnalepa/mbcnn}.

\section{Experimental results}\label{sec:experiments}

The objective of our experiments is three-fold---(\textit{i})~to understand the impact of utilizing parallel branches in the multi-branch architecture trained from the training sets of various sizes, (\textit{ii})~to confront our multi-branch CNNs with other algorithms, and (\textit{iii})~to verify their robustness against noise. Our models were coded in \texttt{Python 3.6} with \texttt{Tensorflow 1.12}. All models were trained using ADAM with the learning rate of $10^{-3}$. The maximum number of epochs was 100, with the early stopping of 15 epochs without an improvement in the loss calculated over a randomly sampled $10\%$ of all training pixels, with a batch size equal to 256. The mean square error (MSE) between the estimated and ground-truth fractional abundances is used as the loss function in the training process.

\begin{figure}[t!]
\pgfplotsset{every axis title/.append style={at={(0.5,0.9)}}}
\center
\begin{tikzpicture}[scale=0.515]		
[font=\footnotesize\sffamily]
\begin{groupplot}[
group style={group size=2 by 3, horizontal sep=0.8cm, vertical sep=0.45cm, group name=my plots},
legend style={at={(0.5,-0.15)},
anchor=north,legend columns=-1},
symbolic x coords={A,B,C,D,E,F,G,H,I,J,K,L,M,N,O,P},
x=0.42cm,xticklabels=\empty,
height=4cm,xtick style={draw=none},
ymajorgrids,y tick label style={/pgf/number format/.cd,fixed,fixed zerofill,precision=2,/tikz/.cd}]
\nextgroupplot[
ylabel={RMSE},
ybar=-5pt,
bar width=0.2cm,
xtick=data,ymax=0.07,ymin=0,title=100\%
]
\draw node[above,xshift=-2mm,yshift=20mm] {\scriptsize$\uparrow$0.20};
\addplot[orange, fill,draw=black, very thick] coordinates {(A,      0.059   )};
\addplot[blue, fill,draw=black, very thick] coordinates {(B,        0.059   )};
\addplot[teal, fill,draw=black, very thick] coordinates {(C,        0.021   )};
\addplot[lime, fill,draw=black, very thick] coordinates {(D,        0.030   )};
\addplot[brown, fill,draw=black, very thick]  coordinates {(E,      0.025   )};
\addplot[gray1, fill,draw=black, very thick]  coordinates {(F,      0.012   )};
\addplot[gray2, fill,draw=black, very thick] coordinates {(G,       0.020   )};
\addplot[gray3, fill,draw=black, very thick] coordinates {(H,       0.013   )};
\addplot[gray4, fill,draw=black, very thick] coordinates {(I,       0.012   )};
\addplot[gray5, fill,draw=black, very thick] coordinates {(J,       0.012   )};
\addplot[gray6, fill,draw=black, very thick] coordinates {(K,       0.012   )};
\addplot[hotmagenta, fill,draw=black, very thick] coordinates {(L,  0.012   )};
\addplot[yellow, fill,draw=black, very thick] coordinates {(M,      0.012   )};
\addplot[red, fill,draw=black, very thick] coordinates {(N,         0.011   )};
\addplot[cyan, fill,draw=black, very thick] coordinates {(O,        0.012   )};
\addplot[green, fill,draw=black, very thick] coordinates {(P,       0.012   )};

\nextgroupplot[
ybar=-5pt,
bar width=0.2cm,
xtick=data,ymax=0.07,ymin=0,title=66\%
]
\draw node[above,xshift=-2mm,yshift=20.5mm] {\scriptsize$\uparrow$0.20};
\addplot[orange, fill,draw=black, very thick] coordinates {(A,      0.062   )};
\addplot[blue, fill,draw=black, very thick] coordinates {(B,        0.062   )};
\addplot[teal, fill,draw=black, very thick] coordinates {(C,        0.022   )};
\addplot[lime, fill,draw=black, very thick] coordinates {(D,        0.033   )};
\addplot[brown, fill,draw=black, very thick]  coordinates {(E,      0.028   )};
\addplot[gray1, fill,draw=black, very thick]  coordinates {(F,      0.014   )};
\addplot[gray2, fill,draw=black, very thick] coordinates {(G,       0.022   )};
\addplot[gray3, fill,draw=black, very thick] coordinates {(H,       0.014   )};
\addplot[gray4, fill,draw=black, very thick] coordinates {(I,       0.014   )};
\addplot[gray5, fill,draw=black, very thick] coordinates {(J,       0.014   )};
\addplot[gray6, fill,draw=black, very thick] coordinates {(K,       0.014   )};
\addplot[hotmagenta, fill,draw=black, very thick] coordinates {(L,  0.014   )};
\addplot[yellow, fill,draw=black, very thick] coordinates {(M,      0.014   )};
\addplot[red, fill,draw=black, very thick] coordinates {(N,         0.013   )};
\addplot[cyan, fill,draw=black, very thick] coordinates {(O,        0.014   )};
\addplot[green, fill,draw=black, very thick] coordinates {(P,       0.013   )};

\nextgroupplot[
ylabel={RMSE},
ybar=-5pt,
bar width=0.2cm,
xtick=data,ymax=0.075,ymin=0,title=33\%
]
\draw node[above,xshift=-2mm,yshift=20.5mm] {\scriptsize$\uparrow$0.20};
\addplot[orange, fill,draw=black, very thick] coordinates {(A,      0.066   )};
\addplot[blue, fill,draw=black, very thick] coordinates {(B,        0.066   )};
\addplot[teal, fill,draw=black, very thick] coordinates {(C,        0.027   )};
\addplot[lime, fill,draw=black, very thick] coordinates {(D,        0.037   )};
\addplot[brown, fill,draw=black, very thick]  coordinates {(E,      0.035   )};
\addplot[gray1, fill,draw=black, very thick]  coordinates {(F,      0.019   )};
\addplot[gray2, fill,draw=black, very thick] coordinates {(G,       0.029   )};
\addplot[gray3, fill,draw=black, very thick] coordinates {(H,       0.020   )};
\addplot[gray4, fill,draw=black, very thick] coordinates {(I,       0.019   )};
\addplot[gray5, fill,draw=black, very thick] coordinates {(J,       0.018   )};
\addplot[gray6, fill,draw=black, very thick] coordinates {(K,       0.020   )};
\addplot[hotmagenta, fill,draw=black, very thick] coordinates {(L,  0.018   )};
\addplot[yellow, fill,draw=black, very thick] coordinates {(M,      0.019   )};
\addplot[red, fill,draw=black, very thick] coordinates {(N,         0.019   )};
\addplot[cyan, fill,draw=black, very thick] coordinates {(O,        0.020   )};
\addplot[green, fill,draw=black, very thick] coordinates {(P,       0.018   )};

\nextgroupplot[
ybar=-5pt,
bar width=0.2cm,
xtick=data,ymax=0.085,ymin=0,title=13\%
]
\draw node[above,xshift=-2mm,yshift=20.5mm] {\scriptsize$\uparrow$0.19};
\addplot[orange, fill,draw=black, very thick] coordinates {(A,      0.074   )};
\addplot[blue, fill,draw=black, very thick] coordinates {(B,        0.074   )};
\addplot[teal, fill,draw=black, very thick] coordinates {(C,        0.036   )};
\addplot[lime, fill,draw=black, very thick] coordinates {(D,        0.043   )};
\addplot[brown, fill,draw=black, very thick]  coordinates {(E,      0.040   )};
\addplot[gray1, fill,draw=black, very thick]  coordinates {(F,      0.029   )};
\addplot[gray2, fill,draw=black, very thick] coordinates {(G,       0.042   )};
\addplot[gray3, fill,draw=black, very thick] coordinates {(H,       0.041   )};
\addplot[gray4, fill,draw=black, very thick] coordinates {(I,       0.029   )};
\addplot[gray5, fill,draw=black, very thick] coordinates {(J,       0.029   )};
\addplot[gray6, fill,draw=black, very thick] coordinates {(K,       0.030   )};
\addplot[hotmagenta, fill,draw=black, very thick] coordinates {(L,  0.029   )};
\addplot[yellow, fill,draw=black, very thick] coordinates {(M,      0.028   )};
\addplot[red, fill,draw=black, very thick] coordinates {(N,         0.030   )};
\addplot[cyan, fill,draw=black, very thick] coordinates {(O,        0.039   )};
\addplot[green, fill,draw=black, very thick] coordinates {(P,       0.027   )};

\nextgroupplot[
ylabel={RMSE},
ybar=-5pt,
bar width=0.2cm,
xtick=data,ymax=0.09,title=6\%
]
\draw node[above,xshift=-2mm,yshift=18.5mm] {\scriptsize$\uparrow$0.20};
\addplot[orange, fill,draw=black, very thick] coordinates {(A,      0.083   )};
\addplot[blue, fill,draw=black, very thick] coordinates {(B,        0.082   )};
\addplot[teal, fill,draw=black, very thick] coordinates {(C,        0.046   )};
\addplot[lime, fill,draw=black, very thick] coordinates {(D,        0.053   )};
\addplot[brown, fill,draw=black, very thick]  coordinates {(E,      0.049   )};
\addplot[gray1, fill,draw=black, very thick]  coordinates {(F,      0.040   )};
\addplot[gray2, fill,draw=black, very thick] coordinates {(G,       0.055   )};
\addplot[gray3, fill,draw=black, very thick] coordinates {(H,       0.030   )};
\addplot[gray4, fill,draw=black, very thick] coordinates {(I,       0.040   )};
\addplot[gray5, fill,draw=black, very thick] coordinates {(J,       0.040   )};
\addplot[gray6, fill,draw=black, very thick] coordinates {(K,       0.041   )};
\addplot[hotmagenta, fill,draw=black, very thick] coordinates {(L,  0.040   )};
\addplot[yellow, fill,draw=black, very thick] coordinates {(M,      0.039   )};
\addplot[red, fill,draw=black, very thick] coordinates {(N,         0.041   )};
\addplot[cyan, fill,draw=black, very thick] coordinates {(O,        0.083   )};
\addplot[green, fill,draw=black, very thick] coordinates {(P,       0.038   )};

\nextgroupplot[
ybar=-5pt,
bar width=0.2cm,
xtick=data,ymax=0.24,ymin=0,title=1\%
]
\draw node[above,xshift=+60mm,yshift=20.5mm] {\scriptsize$\uparrow$0.347};
\addplot[orange, fill,draw=black, very thick] coordinates {(A,      0.214   )};
\addplot[blue, fill,draw=black, very thick] coordinates {(B,        0.110   )};
\addplot[teal, fill,draw=black, very thick] coordinates {(C,        0.090   )};
\addplot[lime, fill,draw=black, very thick] coordinates {(D,        0.092   )};
\addplot[brown, fill,draw=black, very thick]  coordinates {(E,      0.150   )};
\addplot[gray1, fill,draw=black, very thick]  coordinates {(F,      0.082   )};
\addplot[gray2, fill,draw=black, very thick] coordinates {(G,       0.096   )};
\addplot[gray3, fill,draw=black, very thick] coordinates {(H,       0.085   )};
\addplot[gray4, fill,draw=black, very thick] coordinates {(I,       0.081   )};
\addplot[gray5, fill,draw=black, very thick] coordinates {(J,       0.081   )};
\addplot[gray6, fill,draw=black, very thick] coordinates {(K,       0.085   )};
\addplot[hotmagenta, fill,draw=black, very thick] coordinates {(L,  0.080   )};
\addplot[yellow, fill,draw=black, very thick] coordinates {(M,      0.081   )};
\addplot[red, fill,draw=black, very thick] coordinates {(N,         0.085   )};
\addplot[cyan, fill,draw=black, very thick] coordinates {(O,        0.214   )};
\addplot[green, fill,draw=black, very thick] coordinates {(P,       0.142   )};
\end{groupplot}
\end{tikzpicture}
\caption{Overall RMSE (Ur) for different $\TrainingSet$ sizes:
\legendsquare{fill=orange}\,LMM~\cite{heinz1999fully},
\legendsquare{fill=blue}\,SVR~\cite{6638039}
\legendsquare{fill=teal}\,CB-CNN~\cite{cnn_unmixing},
\legendsquare{fill=lime}\,WS-AE~\cite{weaklyae},
\legendsquare{fill=brown}\,UnDIP~\cite{Rasti2021undip},
\legendsquare{fill=gray1}\,MB(1D),
\legendsquare{fill=gray2}\,MB(2D),
\legendsquare{fill=gray3}\,MB(3D),
\legendsquare{fill=gray4}\,MB(1D+2D),
\legendsquare{fill=gray5}\,MB(1D+3D),
\legendsquare{fill=gray6}\,MB(2D+3D),
\legendsquare{fill=hotmagenta}\,MB,
\legendsquare{fill=yellow}\,MB-DR,
\legendsquare{fill=red}\,MB-Res,
\legendsquare{fill=cyan}\,MB-TL,
\legendsquare{fill=green}\,MB-PT. For some methods, we indicate the exact value (out of the Y range) of RMSE above the arrow to maintain readability.}
\label{fig:rmse_urban}
\end{figure}

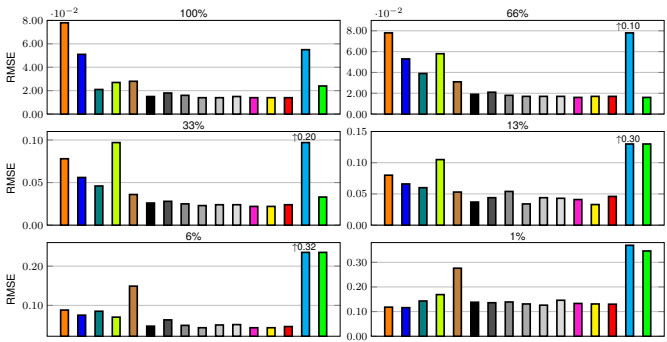
\begin{figure}[b!]
\pgfplotsset{every axis title/.append style={at={(0.5,0.9)}}}
\center
\begin{tikzpicture}[scale=0.515]		
[font=\footnotesize\sffamily]
\begin{groupplot}[
group style={group size=2 by 3, horizontal sep=0.8cm, vertical sep=0.45cm, group name=my plots},
legend style={at={(0.5,-0.15)},
anchor=north,legend columns=-1},
symbolic x coords={A,B,C,D,E,F,G,H,I,J,K,L,M,N,O,P},
x=0.42cm,xticklabels=\empty,
height=4cm,xtick style={draw=none},
ymajorgrids,y tick label style={/pgf/number format/.cd,fixed,fixed zerofill,precision=2,/tikz/.cd}]
\nextgroupplot[
ylabel={RMSE},
ybar=-5pt,
bar width=0.2cm,
xtick=data,ymax=0.08,ymin=0,title=100\%
]
\addplot[orange, fill,draw=black, very thick] coordinates {(A,      0.078   )};
\addplot[blue, fill,draw=black, very thick] coordinates {(B,        0.051   )};
\addplot[teal, fill,draw=black, very thick] coordinates {(C,        0.021   )};
\addplot[lime, fill,draw=black, very thick] coordinates {(D,        0.027   )};
\addplot[brown, fill,draw=black, very thick]  coordinates {(E,      0.028   )};
\addplot[gray1, fill,draw=black, very thick]  coordinates {(F,      0.015   )};
\addplot[gray2, fill,draw=black, very thick] coordinates {(G,       0.018   )};
\addplot[gray3, fill,draw=black, very thick] coordinates {(H,       0.016   )};
\addplot[gray4, fill,draw=black, very thick] coordinates {(I,       0.014   )};
\addplot[gray5, fill,draw=black, very thick] coordinates {(J,       0.014   )};
\addplot[gray6, fill,draw=black, very thick] coordinates {(K,       0.015   )};
\addplot[hotmagenta, fill,draw=black, very thick] coordinates {(L,  0.014   )};
\addplot[yellow, fill,draw=black, very thick] coordinates {(M,      0.014   )};
\addplot[red, fill,draw=black, very thick] coordinates {(N,         0.014   )};
\addplot[cyan, fill,draw=black, very thick] coordinates {(O,        0.055   )};
\addplot[green, fill,draw=black, very thick] coordinates {(P,       0.024   )};

\nextgroupplot[
ybar=-5pt,
bar width=0.2cm,
xtick=data,ymax=0.09,ymin=0,title=66\%
]
\addplot[orange, fill,draw=black, very thick] coordinates {(A,      0.078   )};
\addplot[blue, fill,draw=black, very thick] coordinates {(B,        0.053   )};
\addplot[teal, fill,draw=black, very thick] coordinates {(C,        0.039   )};
\addplot[lime, fill,draw=black, very thick] coordinates {(D,        0.058   )};
\addplot[brown, fill,draw=black, very thick]  coordinates {(E,      0.031   )};
\addplot[gray1, fill,draw=black, very thick]  coordinates {(F,      0.019   )};
\addplot[gray2, fill,draw=black, very thick] coordinates {(G,       0.021   )};
\addplot[gray3, fill,draw=black, very thick] coordinates {(H,       0.018   )};
\addplot[gray4, fill,draw=black, very thick] coordinates {(I,       0.017   )};
\addplot[gray5, fill,draw=black, very thick] coordinates {(J,       0.017   )};
\addplot[gray6, fill,draw=black, very thick] coordinates {(K,       0.017   )};
\addplot[hotmagenta, fill,draw=black, very thick] coordinates {(L,  0.016   )};
\addplot[yellow, fill,draw=black, very thick] coordinates {(M,      0.017   )};
\addplot[red, fill,draw=black, very thick] coordinates {(N,         0.017   )};
\addplot[cyan, fill,draw=black, very thick] coordinates {(O,        0.078   )};
\addplot[green, fill,draw=black, very thick] coordinates {(P,       0.016   )};
\draw node[above,xshift=+60mm,yshift=20.5mm] {\scriptsize$\uparrow$0.10};
\nextgroupplot[
ylabel={RMSE},
ybar=-5pt,
bar width=0.2cm,
xtick=data,ymax=0.11,ymin=0,title=33\%
]
\addplot[orange, fill,draw=black, very thick] coordinates {(A,      0.078   )};
\addplot[blue, fill,draw=black, very thick] coordinates {(B,        0.056   )};
\addplot[teal, fill,draw=black, very thick] coordinates {(C,        0.046   )};
\addplot[lime, fill,draw=black, very thick] coordinates {(D,        0.097   )};
\addplot[brown, fill,draw=black, very thick]  coordinates {(E,      0.036   )};
\addplot[gray1, fill,draw=black, very thick]  coordinates {(F,      0.026   )};
\addplot[gray2, fill,draw=black, very thick] coordinates {(G,       0.028   )};
\addplot[gray3, fill,draw=black, very thick] coordinates {(H,       0.025   )};
\addplot[gray4, fill,draw=black, very thick] coordinates {(I,       0.023   )};
\addplot[gray5, fill,draw=black, very thick] coordinates {(J,       0.024   )};
\addplot[gray6, fill,draw=black, very thick] coordinates {(K,       0.024   )};
\addplot[hotmagenta, fill,draw=black, very thick] coordinates {(L,  0.022   )};
\addplot[yellow, fill,draw=black, very thick] coordinates {(M,      0.022   )};
\addplot[red, fill,draw=black, very thick] coordinates {(N,         0.024   )};
\addplot[cyan, fill,draw=black, very thick] coordinates {(O,        0.097   )};
\addplot[green, fill,draw=black, very thick] coordinates {(P,       0.033   )};
\draw node[above,xshift=+60mm,yshift=20.5mm] {\scriptsize$\uparrow$0.20};
\nextgroupplot[
ybar=-5pt,
bar width=0.2cm,
xtick=data,ymax=0.15,ymin=0,title=13\%
]
\addplot[orange, fill,draw=black, very thick] coordinates {(A,      0.080   )};
\addplot[blue, fill,draw=black, very thick] coordinates {(B,        0.066   )};
\addplot[teal, fill,draw=black, very thick] coordinates {(C,        0.060   )};
\addplot[lime, fill,draw=black, very thick] coordinates {(D,        0.105   )};
\addplot[brown, fill,draw=black, very thick]  coordinates {(E,      0.053   )};
\addplot[gray1, fill,draw=black, very thick]  coordinates {(F,      0.037   )};
\addplot[gray2, fill,draw=black, very thick] coordinates {(G,       0.044   )};
\addplot[gray3, fill,draw=black, very thick] coordinates {(H,       0.054   )};
\addplot[gray4, fill,draw=black, very thick] coordinates {(I,       0.034   )};
\addplot[gray5, fill,draw=black, very thick] coordinates {(J,       0.044   )};
\addplot[gray6, fill,draw=black, very thick] coordinates {(K,       0.043   )};
\addplot[hotmagenta, fill,draw=black, very thick] coordinates {(L,  0.041   )};
\addplot[yellow, fill,draw=black, very thick] coordinates {(M,      0.033   )};
\addplot[red, fill,draw=black, very thick] coordinates {(N,         0.046   )};
\addplot[cyan, fill,draw=black, very thick] coordinates {(O,        0.130   )};
\addplot[green, fill,draw=black, very thick] coordinates {(P,       0.130   )};
\draw node[above,xshift=+60mm,yshift=20mm] {\scriptsize$\uparrow$0.30};
\nextgroupplot[
ylabel={RMSE},
ybar=-5pt,
bar width=0.2cm,
xtick=data,ymax=0.26,title=6\%
]
\addplot[orange, fill,draw=black, very thick] coordinates {(A,      0.088   )};
\addplot[blue, fill,draw=black, very thick] coordinates {(B,        0.075   )};
\addplot[teal, fill,draw=black, very thick] coordinates {(C,        0.085   )};
\addplot[lime, fill,draw=black, very thick] coordinates {(D,        0.070   )};
\addplot[brown, fill,draw=black, very thick]  coordinates {(E,      0.149   )};
\addplot[gray1, fill,draw=black, very thick]  coordinates {(F,      0.047   )};
\addplot[gray2, fill,draw=black, very thick] coordinates {(G,       0.063   )};
\addplot[gray3, fill,draw=black, very thick] coordinates {(H,       0.049   )};
\addplot[gray4, fill,draw=black, very thick] coordinates {(I,       0.043   )};
\addplot[gray5, fill,draw=black, very thick] coordinates {(J,       0.050   )};
\addplot[gray6, fill,draw=black, very thick] coordinates {(K,       0.051   )};
\addplot[hotmagenta, fill,draw=black, very thick] coordinates {(L,  0.043   )};
\addplot[yellow, fill,draw=black, very thick] coordinates {(M,      0.043   )};
\addplot[red, fill,draw=black, very thick] coordinates {(N,         0.046   )};
\addplot[cyan, fill,draw=black, very thick] coordinates {(O,        0.235   )};
\addplot[green, fill,draw=black, very thick] coordinates {(P,       0.235   )};
\draw node[above,xshift=+60mm,yshift=18.5mm] {\scriptsize$\uparrow$0.32};
\nextgroupplot[
ybar=-5pt,
bar width=0.2cm,
xtick=data,ymax=0.38,ymin=0,title=1\%
]
\addplot[orange, fill,draw=black, very thick] coordinates {(A,      0.118   )};
\addplot[blue, fill,draw=black, very thick] coordinates {(B,        0.116   )};
\addplot[teal, fill,draw=black, very thick] coordinates {(C,        0.143   )};
\addplot[lime, fill,draw=black, very thick] coordinates {(D,        0.169   )};
\addplot[brown, fill,draw=black, very thick]  coordinates {(E,      0.276   )};
\addplot[gray1, fill,draw=black, very thick]  coordinates {(F,      0.138   )};
\addplot[gray2, fill,draw=black, very thick] coordinates {(G,       0.136   )};
\addplot[gray3, fill,draw=black, very thick] coordinates {(H,       0.139   )};
\addplot[gray4, fill,draw=black, very thick] coordinates {(I,       0.131   )};
\addplot[gray5, fill,draw=black, very thick] coordinates {(J,       0.126   )};
\addplot[gray6, fill,draw=black, very thick] coordinates {(K,       0.146   )};
\addplot[hotmagenta, fill,draw=black, very thick] coordinates {(L,  0.133   )};
\addplot[yellow, fill,draw=black, very thick] coordinates {(M,      0.131   )};
\addplot[red, fill,draw=black, very thick] coordinates {(N,         0.130   )};
\addplot[cyan, fill,draw=black, very thick] coordinates {(O,        0.369   )};
\addplot[green, fill,draw=black, very thick] coordinates {(P,       0.346   )};

\end{groupplot}
\end{tikzpicture}
\caption{Overall RMSE (JR) for different $\TrainingSet$ sizes:
\legendsquare{fill=orange}\,LMM~\cite{heinz1999fully},
\legendsquare{fill=blue}\,SVR~\cite{6638039}
\legendsquare{fill=teal}\,CB-CNN~\cite{cnn_unmixing},
\legendsquare{fill=lime}\,WS-AE~\cite{weaklyae},
\legendsquare{fill=brown}\,UnDIP~\cite{Rasti2021undip},
\legendsquare{fill=gray1}\,MB(1D),
\legendsquare{fill=gray2}\,MB(2D),
\legendsquare{fill=gray3}\,MB(3D),
\legendsquare{fill=gray4}\,MB(1D+2D),
\legendsquare{fill=gray5}\,MB(1D+3D),
\legendsquare{fill=gray6}\,MB(2D+3D),
\legendsquare{fill=hotmagenta}\,MB,
\legendsquare{fill=yellow}\,MB-DR,
\legendsquare{fill=red}\,MB-Res,
\legendsquare{fill=cyan}\,MB-TL,
\legendsquare{fill=green}\,MB-PT.}
\label{fig:rmse_jasper_ridge}
\end{figure}

We confront our multi-branch CNN with LMM~\cite{heinz1999fully}, SVR with one regression model fitted per target for the multi-target unmixing~\cite{6638039}, and recent approaches such as the cube-based variant of CNN (CB-CNN) that extracts the spectral-spatial features~\cite{cnn_unmixing}, the WS-AE architecture~\cite{weaklyae}, and the UnDIP architecture~\cite{Rasti2021undip}. Apart from LMM, each algorithm operates on a 3D patch of size $3\times 3\times \AllBands$, and all of the methods were trained in a supervised way, utilizing the ground-truth abundances (therefore, all of them perform the estimation of endmember fractional abundances, and they do \textit{not} automatically determine the set of all endmembers in the scene). In the ablation study, we analyze all combinations of the 1D, 2D, and 3D branches in MB---the MB variants which include a single branch are MB(1D), MB(2D), and MB(3D), whereas those with two parallel branches encompass MB(1D+2D), MB(1D+3D), and MB(2D+3D), together with the impact of the training set size on their capabilities. Additionally, we generated the contaminated test sets with the white zero-mean Gaussian noise added to the original test data (the signal-to-noise ratio of 20, 30, 40, and 50 dB), to verify the robustness of the models against noise. As the unmixing quality metrics, we capture the root mean square error (RMSE), and the root mean square abundance angle distance (rmsAAD)~\cite{8984691}.

%${\rm RMSE}=\sqrt{{\sum_{i=1}^{\left|\TestSet\right|}(\vect{a}_i-\vect{\hat{a}}_i)^2}/{\left|\TestSet\right|}}$,
%where $\left|\TestSet\right|$ is the number of samples in the test set $\TestSet$, and $\vect{a}$ and $\vect{\hat{a}}$ are the observed and estimated abundance vectors, and the root mean square abundance angle distance ${\rm rmsAAD}=\sqrt{{\sum_{i=1}^{\left|\TestSet\right|}\arccos(\frac{\vect{a}^{\top}_{i}\vect{\hat{a}}_i}{\Vert{\vect{a}_i}\Vert\Vert\vect{\hat{a}}_i\Vert})^2}/{\left|\TestSet\right|}}$.

\begin{table}[t!]
\caption{The results of the two-tailed Wilcoxon tests ($p<0.05$)---we present the number of cases (for each training set size, out of 3 HU sets) in which the confronted variants lead to obtaining statistically the same results as those by MB. We boldface the entries, in which MB obtained the statistically significantly better results for all sets.}
\label{tab:wilcoxon}
\scriptsize
\centering
\renewcommand{\tabcolsep}{1.9mm}
\begin{tabular}{lccccccr}
\Xhline{2\arrayrulewidth}
Compared with $\downarrow$&
  \textbf{100\%} & \textbf{66\%} & \textbf{33\%} & \textbf{13\%} & \textbf{6\%} & \textbf{1\%} & Total\\ \hline
  MB(1D) & \textbf{0/3} & \textbf{0/3} & \textbf{0/3} & 1/3 & 2/3 & 2/3 & 5/18\\
  MB(2D) & \textbf{0/3} & \textbf{0/3} & \textbf{0/3} & \textbf{0/3} & \textbf{0/3} & 1/3 & 1/18\\
  MB(3D) & \textbf{0/3} & \textbf{0/3} & \textbf{0/3} & 1/3 & 1/3 & 3/3 & 5/18\\
  MB(1D+2D) & 1/3 & \textbf{0/3} & \textbf{0/3} & 2/3 & 2/3 & 2/3 & 7/18\\
  MB(1D+3D) & 3/3 & 1/3 & 1/3 & 2/3 & 3/3 & 3/3 & 13/18\\
  MB(2D+3D) & 1/3 & \textbf{0/3} & \textbf{0/3} & \textbf{0/3} & 2/3 & 2/3 & 5/18\\
  \hline
  Total$\rightarrow$ & 5/18 & 1/18 & 1/18 & 6/18 & 10/18 & 13/18 & 36/108\\
\Xhline{2\arrayrulewidth}           
\end{tabular}
\end{table} 

% \usepackage[normalem]{ulem}
\begin{table*}[ht!]
\caption{The RMSE and rmsAAD metrics obtained for all investigated algorithms, averaged across all (30) test sets elaborated for each benchmark dataset (Sa, Ur, and JR). The best result for each $\TrainingSet$ size is boldfaced, whereas the second best is underlined. The background of the globally best result (across all training set sizes) is rendered in green. }\label{tab:detailed_results}
\centering
\setlength{\tabcolsep}{2.9mm}
\scalebox{0.9}{
\begin{tabular}{rrrrrrrrrrrrrrrr}
\Xhline{2\arrayrulewidth}
\multicolumn{1}{c}{}              & \multicolumn{7}{c}{\textbf{RMSE}}                                                                                                                                 && \multicolumn{7}{c}{\textbf{rmsAAD}}                                                                                                                   \\
\cline{2-8} \cline{10-16}
%\Xhline{1.5\arrayrulewidth}
\textbf{Models}                         & \textbf{100\%} & \textbf{66\%} & \textbf{33\%} & \textbf{13\%} & \textbf{6\%} & \textbf{1\%} & \textbf{Mean}              && \textbf{100\%} & \textbf{66\%} & \textbf{33\%} & \textbf{13\%} & \textbf{6\%} & \textbf{1\%} & \textbf{Mean}  \\ \Xhline{1.5\arrayrulewidth}
LMM~\cite{heinz1999fully}         & 0.230          & 0.230                 & 0.230                 & 0.231                 & 0.234                & 0.250                & 0.234 && 0.588          & 0.588                 & 0.588                 & 0.590                 & 0.598                & 0.645                & 0.599          \\
SVR~\cite{6638039}         & 0.054          & 0.057                 & 0.061                 & 0.070                 & 0.079                & 0.113                & 0.072 && 0.132          & 0.138                 & 0.151                 & 0.175                 & 0.199                & 0.292                & 0.181          \\
CB-CNN~\cite{cnn_unmixing}      & 0.020          & 0.027                 & 0.032                 & 0.043                 & 0.058                & 0.103                & 0.047 && 0.053          & 0.071                 & 0.084                 & 0.114                 & 0.153                & 0.276                & 0.125          \\
WS-AE~\cite{weaklyae}        & 0.026          & 0.038                 & 0.054                 & 0.061                 & 0.055                & 0.129                & 0.061 && 0.071          & 0.099                 & 0.137                 & 0.156                 & 0.146                & 0.335                & 0.157          \\
UnDIP~\cite{Rasti2021undip} & 0.030	&0.035	&0.050	&0.084	&0.127	&0.243	&0.095&&0.075	&0.085	&0.121	&0.191	&0.304	&0.615	&0.232\\
\cline{1-8} \cline{10-16}
MB(1D)    & {\ul 0.013}          & 0.016                 & 0.021                 & {\ul0.031}                 & 0.041                & 0.108                & 0.038 && 0.034          & 0.041                 & 0.056                 & 0.083                 & 0.111                & 0.282                & 0.101          \\

MB(2D)    & 0.018          & 0.020                 & 0.028                 & 0.041                 & 0.054                & 0.116                & 0.046 && 0.049          & 0.056                 & 0.077                 & 0.115                 & 0.151                & 0.306                & 0.126          \\

MB(3D)    & 0.014          &{\ul 0.015}                 & 0.021                 & 0.041                 & \textbf{0.039}                & 0.105                & 0.039 && 0.036          & 0.042                 & 0.056                 & 0.111                 & \textbf{0.103}                & 0.280                & 0.105          \\
\cline{1-8} \cline{10-16}
MB(1D+2D) & \cellcolor{green!25}{\textbf{0.012}}          & \textbf{0.014}                 & {\ul0.020}                 & \textbf{0.029}                 & {\ul0.040}                & 0.104                & 0.037  && 0.033          & 0.039               & {\ul0.053}                 & {\ul0.080}                 & 0.109                & 0.272                & 0.098          \\

MB(1D+3D) & \cellcolor{green!25}\textbf{0.012}          &{\ul 0.015}                 & {\ul0.020}                 & 0.032                 & 0.041                & \textbf{0.095}                & {\ul 0.036}                && \cellcolor{green!25}\textbf{0.031}          & {\ul0.038}                 & {\ul0.053}                 & 0.087                 & 0.113                & \textbf{0.253}                & 0.096          \\

MB(2D+3D)                      & 0.013          & \textbf{0.014}                 & 0.021                 & 0.033                 & 0.043                & 0.111                & 0.039                      && 0.034          & 0.039                 & 0.055                 & 0.090                 & 0.118                & 0.293                & 0.105          \\
\cline{1-8} \cline{10-16}
\textbf{MB}                       & \cellcolor{green!25}\textbf{0.012}          & \textbf{0.014}              & \textbf{0.019}                 & {\ul0.031}                 & \textbf{0.039}                & \textbf{0.095}                & \textbf{0.035}             && \cellcolor{green!25}\textbf{0.031}          & \textbf{0.037}                 & \textbf{0.051}                 & 0.084                 & 0.106                & {\ul0.255}                & \textbf{0.094} \\

\textbf{MB-DR}                    & \cellcolor{green!25}\textbf{0.012}          & \textbf{0.014}            & \textbf{0.019}                 & \textbf{0.029}                 & \textbf{0.039}                & {\ul0.100}                & \textbf{0.035}             && {\ul0.032}          & 0.039                 & \textbf{0.051}                 & \textbf{0.077}                 & {\ul0.105}                & 0.263                & {\ul 0.095}    \\

\textbf{MB-Res}                   & \cellcolor{green!25}\textbf{0.012}          & \textbf{0.014}            & {\ul0.020}                 & 0.034                 & 0.041                & {\ul0.100}                & 0.037                      && \cellcolor{green!25}\textbf{0.031}          & {\ul0.038}                 & {\ul0.053}                 & 0.090                 & 0.111                & 0.263                & 0.098          \\

\textbf{MB-TL}                    & 0.027          & 0.048                 & 0.104                 & 0.163                 & 0.213                & 0.342                & 0.150                      && 0.070          & 0.124                 & 0.254                 & 0.398                 & 0.528                & 0.871                & 0.374          \\

\textbf{MB-PT}                    & 0.015          & 0.018                 & 0.023                 & 0.065                 & 0.107                & 0.217                & 0.074                      && 0.038          & 0.045                 & 0.059                 & 0.164                 & 0.270                & 0.545                & 0.187         \\\Xhline{2\arrayrulewidth}
\end{tabular}}
\end{table*}

% \usepackage[normalem]{ulem}
\begin{table*}[ht!]
\caption{The ranking (RMSE and rmsAAD) obtained by all algorithms, averaged across all sets (Sa, Ur, and JR). The best ranking for each $\TrainingSet$ size is boldfaced, whereas the second best is underlined. For the ranking obtained for each set, see supplementary material.}\label{tab:ranking}
\centering
\renewcommand{\tabcolsep}{2.3mm}
\scalebox{0.9}{
\begin{tabular}{rrrrrrrrrrrrrrrr}
\Xhline{2\arrayrulewidth}
\multicolumn{1}{c}{}              & \multicolumn{7}{c}{\textbf{RMSE}}                                                                                                                                 & \multicolumn{7}{c}{\textbf{rmsAAD}}                                                                                                                   \\
\cline{2-8} \cline{10-16}
\textbf{Train. size $\rightarrow$}                         & \textbf{100\%} & \textbf{66\%} & \textbf{33\%} & \textbf{13\%} & \textbf{6\%} & \textbf{1\%} & \textbf{Mean}              && \textbf{100\%} & \textbf{66\%} & \textbf{33\%} & \textbf{13\%} & \textbf{6\%} & \textbf{1\%} & \textbf{Mean}  \\\Xhline{1.5\arrayrulewidth}
LMM~\cite{heinz1999fully}             & 16.000&	15.667&	15.333&	15.000&	15.000&	11.000&	14.667&& 16.000&	15.667&	15.333	&15.000	&15.000	&11.000	&14.667\\
SVR~\cite{6638039}             & 14.667	&14.333&	13.667&	13.333&	12.667&	7.667&	12.722&& 14.667&	14.333&	13.667&	13.333&	12.667&	7.667&	12.722\\
CB-CNN~\cite{cnn_unmixing}         & 11.333&	11.333&	11.000&	9.667&	10.333&	7.333&	10.167&&  11.333&	11.000&	11.000&	9.667&	10.333&	7.333&	10.111 \\
WS-AE~\cite{weaklyae}         & 13.000&	13.000&	13.667&	12.667&	10.667&	11.667&	12.444&& 13.000&	13.000&	13.333&	12.333&	10.667&	11.667&	12.333\\
UnDIP~\cite{Rasti2021undip} &13.333 &	12.667 &	12.667 &	11.667 &	13.000 &	14.000 &	12.889  &&13.333 &	12.667 &	12.667 &	11.333 &	13.000 &	14.333 &	12.889\\ \cline{1-8} \cline{10-16}
MB(1D)       & 6.833&	7.833&	7.333&	4.500&	5.500&	7.667&	6.611&& 7.333&	7.833&	7.833&	4.500&	4.667&	7.333&	6.583\\
MB(2D)       &10.333&	10.000&	10.667&	10.167&	11.000&	10.000&	10.361&& 10.333&	10.333&	11.000&	11.000&	11.000&	10.000&	10.611\\
MB(3D)       & 9.000&	7.500&	8.500&	10.000&	5.000&	7.500&	7.917&& 8.500&	8.000&	8.333&	10.000&	5.000&	7.500&	7.889 \\ \cline{1-8} \cline{10-16}
MB(1D+2D)    & {\ul4.333}&	5.167&	4.667&	{\ul3.667}&	4.833&	5.167&	4.639&& 6.500&	6.333&	5.000&	3.833&	4.667&	5.167&	5.250 \\
MB(1D+3D)    &{\ul4.333}&	6.333&	3.833&	4.167&	5.000&	\textbf{2.833}&	4.417 &&3.167&	4.667&	4.333&	{\ul3.333}&	5.000&	{\ul3.000}&	3.917 \\
MB(2D+3D)    &6.500	&5.167	&7.500&	6.500&	7.667&	8.833&	7.028 &&6.500&	5.833&	6.667&	6.667&	8.000&	9.500&	7.194          \\\cline{1-8} \cline{10-16}
\textbf{MB}     & {\ul4.333}	&\textbf{2.000}&	\textbf{1.667}&	{\ul3.667}&	\textbf{2.333}&	{\ul3.000}&	\textbf{2.833}&& {\ul2.833}&	\textbf{2.000}&	\textbf{1.500}&	4.000&	{\ul3.333}&	\textbf{2.833}&	\textbf{2.750}\\
\textbf{MB-DR}  &{\ul4.333}&	5.833&	4.333&	\textbf{2.167}&	{\ul2.667}&	4.667&	{\ul4.000} &&4.500&	5.000&	{\ul3.333}&	\textbf{2.000}	&\textbf{2.333}&	4.667	&{\ul3.639}\\
\textbf{MB-Res} & \textbf{2.667}&	{\ul3.500}&	{\ul3.333}&	6.167&	5.333&	5.333&	4.389&& \textbf{2.667}&	{\ul3.000}&	4.000&	6.333	&5.000	&5.000&	4.333\\
\textbf{MB-TL}  &10.333	&11.000	&13.000	&13.333	&15.333	&15.667	&13.111&& 10.667&	11.667&	13.667	&13.667	&15.333	&15.333	&13.389   \\
\textbf{MB-PT}  &4.667&	4.667	&4.833&	9.333	&9.667	&13.667&	7.806 &&4.667	&4.667	&4.333	&9.000&	10.000&	13.667&	7.722\\\Xhline{2\arrayrulewidth}
\end{tabular}}
\end{table*}

We focus on three benchmarks: {Samson} (Sa, $95\times95$, 156 bands), {Urban} (Ur, $207\times307$, 162 bands) and {Jasper Ridge} (JR, $100\times100$, 198 bands)~\cite{6923488,4694061}. The Sa set incorporates three endmembers: \#1 Soil, \#2 Tree and \#3 Water, for Ur we have six endmembers: \#1 Asphalt, \#2 Grass, \#3 Tree, \#4 Roof, \#5 Metal, and \#6 Dirt, whereas for JR, there are four endmembers: \#1 Road, \#2 Water, \#3 Soil, and \#4 Tree. We perform 30-fold Monte Carlo cross-validation, and sample 30 test sets that do \emph{not change} with the change of the training set ($\TrainingSet$) size---we always report the results obtained for the unseen test sets ($\TestSet$), whose size is kept constant as suggested in~\cite{8432512}, and equals 3025, 47249, and 2500 for Sa, Ur, and JR. For Sa, Ur and JR, we have $6\cdot10^3$, $47\cdot 10^3$ and $7.5\cdot 10^3$ training pixels in total. To verify the impact of the training set sizes, we use the subsets of the full $\TrainingSet$'s: $\{1, 6, 13, 33, 66\}\%$, therefore sample the reduced sets of $\{60, 360, 7.8\cdot10^2, 1.98\cdot 10^3, 3.96\cdot 10^3\}$, $\{470, 2.8\cdot 10^3, 6.1\cdot 10^3, 15.5\cdot 10^3, 31\cdot 10^3\}$ and $\{75, 500, 10^3, 2.5\cdot 10^3, 5\cdot 10^3\}$ pixels for Sa, Ur and JR.

The results obtained for all datasets, sizes of $\TrainingSet$'s and algorithms (Figs.~\ref{fig:rmse_urban}--\ref{fig:rmse_jasper_ridge}; for RMSE elaborated for Sa, which is consistent with Ur, and the rmsAAD plots see the supplementary material) indicate that using all branches in MB leads to statistically significantly better unmixing in the majority (72/108) of cases, as confirmed by the Wilcoxon tests (Table~\ref{tab:wilcoxon} for RMSE, $p<0.05$). Here, we report the results obtained for confronting MB with its variants that utilize a subset of all branches (the Wilcoxon tests for all algorithms are at \url{https://gitlab.com/jnalepa/mbhu}). The experiments show that using the 1D and 3D branches brings the largest benefits, and further fusing them in the multi-branch processing leads to the improved unmixing. Fusing the 2D and 3D branches does not significantly enhance the capabilities of the variant utilizing the 3D branch only---both MB(3D) and MB(2D+3D) led to statistically identical results as MB in 5/18 cases. It can be attributed to the fact that both branches capture spectral-spatial characteristics, hence they may lead to similar features. On the other hand, notable improvements are observed once spectral and spectral-spatial branches are fused. The least visible differences are manifested for the smallest training sets---it indicates that the performance of CNNs is becoming saturated for such small samples, and cannot be further improved without capturing or synthesizing more training examples~\cite{audebert_generative_2018}.

\begin{figure}[b!]
    \centering
    %~\\[-3mm]
    %\hspace*{-0.5cm}
    \includegraphics[width=1\columnwidth]{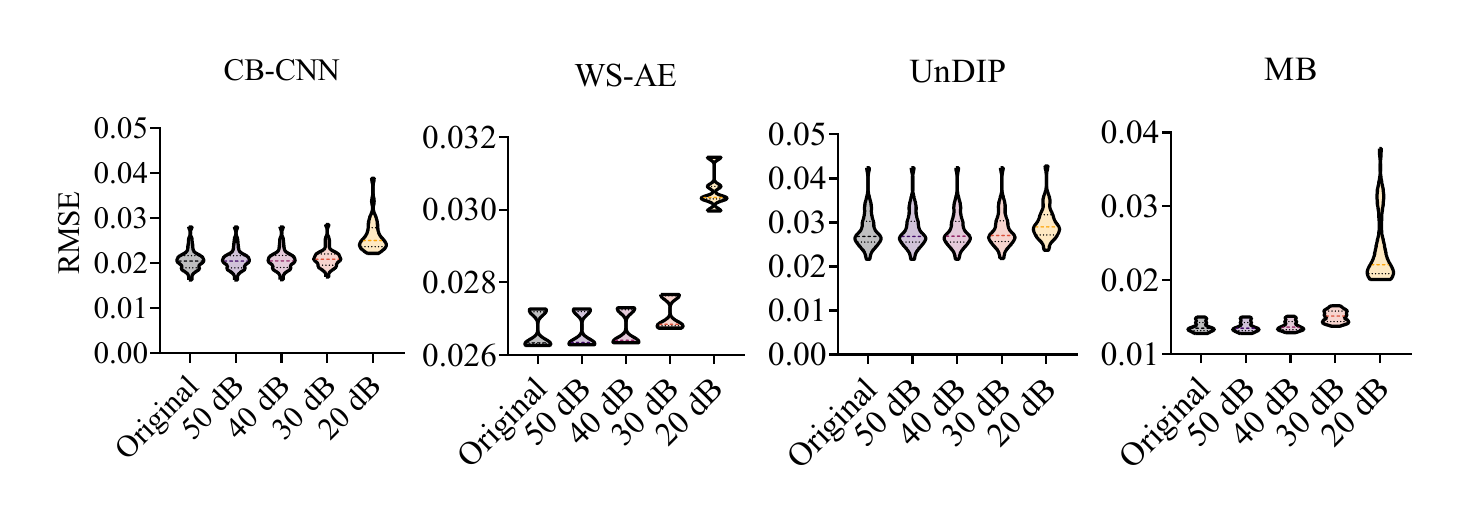}\\
    \caption{The impact of the white zero-mean Gaussian noise added to the original test data (Jasper Ridge) on RMSE obtained by the selected models trained from the full training set.}
    \label{fig:noise}
\end{figure}

%noise_selected_nop_cut.png

The MB CNNs consistently outperform other techniques in all sets and training set sizes, also when averaged across all sizes of $\TrainingSet$ (Table~\ref{tab:detailed_results}), with MB and MB-DR being the Top-2 methods for RMSE and rmsAAD, according to the ranking tests (Table~\ref{tab:ranking}). It indicates that capturing spectral and spectral-spatial features through automated representation learning in parallel branches, and effectively fusing them leads to more precise fractional abundance estimation. For virtually all techniques, increasing $\TrainingSet$'s significantly improves the HU quality over $\TestSet$'s which remained unchanged during the experimentation. The results also indicate that decoupling the feature extraction branches from each other in MB-TL adversely affects the generalization of the model, as it gave the worst performance in all cases when compared to other MB CNNs. The detailed results aggregated for all separate executions across all scenarios gathered in Table~\ref{tab:detailed_results} confirm that the multi-branch architectures outperform other methods investigated in this study. Finally, the visualizations of the abundance maps included in the supplementary material confirm the unmixing abilities of the investigated techniques, and their quality strictly corresponds to the quantitative metrics (RMSE and rmsAAD).

\begin{table}[b!]
\scriptsize
\centering
\setlength{\tabcolsep}{3.5mm}
\caption{The impact of the patch size on RMSE (on Jasper Ridge), quantified as $\Delta_{\rm RMSE}={\rm RMSE}^{k\times k}-{\rm RMSE}^{3\times 3}$ ($k=\{5, 7, 9\}$) obtained using MB trained from the training sets of various sizes.}\label{tab:impact_patch}
\begin{tabular}{cccccc}
\Xhline{2\arrayrulewidth}
& RMSE & & \multicolumn{3}{c}{$\Delta_{\rm RMSE}$}\\
\cline{2-2} \cline{4-6}
\multicolumn{1}{c}{\textbf{Train. size}} & \multicolumn{1}{c}{$3\times   3$} && \multicolumn{1}{c}{$5\times 5$} & \multicolumn{1}{c}{$7\times7$} & \multicolumn{1}{c}{$9\times 9$} \\
\hline
100\%                          & 0.014                                          && 0.002                                          & 0.002                                          & 0.003                                          \\
66\%                           & 0.016                                          && 0.002                                          & 0.002                                          & 0.003                                          \\
33\%                           & 0.022                                          && 0.103                                          & 0.005                                          & 0.008                                          \\
13\%                           & 0.032                                          && 0.003                                          & 0.009                                          & 0.017                                          \\
6\%                            & 0.042                                          && 0.004                                          & 0.010                                          & 0.022                                          \\
1\%                            & 0.105                                          && 0.028                                          & 0.055                                          & 0.038                                         \\
\Xhline{2\arrayrulewidth}
\end{tabular}
\end{table}

To investigate the robustness of the models against noise, we render MB, together with the three best-performing algorithms from the literature in Fig.~\ref{fig:noise} (all 30 $\TestSet$'s were independently contaminated with noise for JR). Here, although UnDIP presents the best robustness against additive white Gaussian noise, both UnDIP and MP led to statistically same results for the uncontaminated $\TestSet$'s and those with the signal-to-noise of 50 dB ($p<0.01$; for all algorithms, see supplementary material). Additionally, in Table~\ref{tab:impact_patch}, we quantify the influence of different input patch sizes on RMSE obtained for JR using MB (executed for five test sets). The results show that analyzing too large pixel neighborhoods leads to less stable models---for the $5\times 5$, $7\times 7$ and $9\times 9$ patches, RMSE's standard deviation amounted to 0.220, 0.209, and 0.215, whereas for $3\times 3$: 0.002. This may have been caused by too large pixel sizes (GSD) which negatively impacted the unmixing process of the central pixel because of capturing different materials.

\begin{figure}[t!]
\center
\hspace*{-0.3cm}
\begin{tikzpicture}[scale=0.52]		
[font=\footnotesize\sffamily]
\begin{groupplot}[
group style={group size=2 by 3, horizontal sep=0.8cm, vertical sep=0.7cm, group name=my plots},
legend style={at={(0.5,-0.15)},
anchor=north,legend columns=-1},
symbolic x coords={A,B,C,D,E,F,G,H,I,J,K,L,M,N,O,P},
x=0.42cm,xticklabels=\empty,
ymode=log,
height=4cm,xtick style={draw=none},
ymajorgrids,ymin=1,
]
\nextgroupplot[
ylabel={Time (s)},
ybar=-5pt,
bar width=0.2cm,
xtick=data,
]
\addplot[orange, fill,draw=black, very thick] coordinates {(A,9.087)};
\addplot[blue, fill,draw=black, very thick] coordinates {(B,3712.361)};
\addplot[teal, fill,draw=black, very thick] coordinates {(C,1429.302)};
\addplot[lime, fill,draw=black, very thick] coordinates {(D,20171.006)};
\addplot[brown, fill,draw=black, very thick]  coordinates {(E,123.856)};
\addplot[gray1, fill,draw=black, very thick]  coordinates {(F,143.371)};
\addplot[gray2, fill,draw=black, very thick] coordinates {(G,141.952)};
\addplot[gray3, fill,draw=black, very thick] coordinates {(H,2440.398)};
\addplot[gray4, fill,draw=black, very thick] coordinates {(I,157.427)};
\addplot[gray5, fill,draw=black, very thick] coordinates {(J,2862.119)};
\addplot[gray6, fill,draw=black, very thick] coordinates {(K,2761.493)};
\addplot[hotmagenta, fill,draw=black, very thick] coordinates {(L,2631.622)};
\addplot[yellow, fill,draw=black, very thick] coordinates {(M,2490.854)};
\addplot[red, fill,draw=black, very thick] coordinates {(N,2535.475)};
\addplot[cyan, fill,draw=black, very thick] coordinates {(O,3720.475)};
\addplot[green, fill,draw=black, very thick] coordinates {(P,5133.646)};

\nextgroupplot[
ybar=-5pt,
bar width=0.2cm,
xtick=data,
]
\addplot[orange, fill,draw=black, very thick] coordinates {(A,7.199)};
\addplot[blue, fill,draw=black, very thick] coordinates {(B,784.861)};
\addplot[teal, fill,draw=black, very thick] coordinates {(C,1032.489)};
\addplot[lime, fill,draw=black, very thick] coordinates {(D,12798.948)};
\addplot[brown, fill,draw=black, very thick]  coordinates {(E,83.138)};
\addplot[gray1, fill,draw=black, very thick]  coordinates {(F,77.640)};
\addplot[gray2, fill,draw=black, very thick] coordinates {(G,78.569)};
\addplot[gray3, fill,draw=black, very thick] coordinates {(H,1851.842)};
\addplot[gray4, fill,draw=black, very thick] coordinates {(I,107.322)};
\addplot[gray5, fill,draw=black, very thick] coordinates {(J,1915.756)};
\addplot[gray6, fill,draw=black, very thick] coordinates {(K,1911.868)};
\addplot[hotmagenta, fill,draw=black, very thick] coordinates {(L,1806.609)};
\addplot[yellow, fill,draw=black, very thick] coordinates {(M,1634.714)};
\addplot[red, fill,draw=black, very thick] coordinates {(N,1730.021)};
\addplot[cyan, fill,draw=black, very thick] coordinates {(O,2420.739)};
\addplot[green, fill,draw=black, very thick] coordinates {(P,3442.744)};

%Salinas
\nextgroupplot[
ylabel={Time (s)},
ybar=-5pt,
bar width=0.2cm,
xtick=data,
]
\addplot[orange, fill,draw=black, very thick] coordinates {(A,7.171)};
\addplot[blue, fill,draw=black, very thick] coordinates {(B,224.143)};
\addplot[teal, fill,draw=black, very thick] coordinates {(C,580.175)};
\addplot[lime, fill,draw=black, very thick] coordinates {(D,5768.140)};
\addplot[brown, fill,draw=black, very thick]  coordinates {(E,40.665)};
\addplot[gray1, fill,draw=black, very thick]  coordinates {(F,39.264)};
\addplot[gray2, fill,draw=black, very thick] coordinates {(G,41.054)};
\addplot[gray3, fill,draw=black, very thick] coordinates {(H,932.516)};
\addplot[gray4, fill,draw=black, very thick] coordinates {(I,54.017)};
\addplot[gray5, fill,draw=black, very thick] coordinates {(J,983.344)};
\addplot[gray6, fill,draw=black, very thick] coordinates {(K,951.281)};
\addplot[hotmagenta, fill,draw=black, very thick] coordinates {(L,916.081)};
\addplot[yellow, fill,draw=black, very thick] coordinates {(M,847.862)};
\addplot[red, fill,draw=black, very thick] coordinates {(N,872.097)};
\addplot[cyan, fill,draw=black, very thick] coordinates {(O,1193.685)};
\addplot[green, fill,draw=black, very thick] coordinates {(P,1665.454)};

\nextgroupplot[
ybar=-5pt,
bar width=0.2cm,
xtick=data,
]
\addplot[orange, fill,draw=black, very thick] coordinates {(A,6.442)};
\addplot[blue, fill,draw=black, very thick] coordinates {(B,47.058)};
\addplot[teal, fill,draw=black, very thick] coordinates {(C,236.723)};
\addplot[lime, fill,draw=black, very thick] coordinates {(D,2328.239)};
\addplot[brown, fill,draw=black, very thick]  coordinates {(E,20.376)};
\addplot[gray1, fill,draw=black, very thick]  coordinates {(F,16.149)};
\addplot[gray2, fill,draw=black, very thick] coordinates {(G,16.942)};
\addplot[gray3, fill,draw=black, very thick] coordinates {(H,366.689)};
\addplot[gray4, fill,draw=black, very thick] coordinates {(I,21.903)};
\addplot[gray5, fill,draw=black, very thick] coordinates {(J,376.090)};
\addplot[gray6, fill,draw=black, very thick] coordinates {(K,353.083)};
\addplot[hotmagenta, fill,draw=black, very thick] coordinates {(L,348.649)};
\addplot[yellow, fill,draw=black, very thick] coordinates {(M,353.923)};
\addplot[red, fill,draw=black, very thick] coordinates {(N,354.910)};
\addplot[cyan, fill,draw=black, very thick] coordinates {(O,460.832)};
\addplot[green, fill,draw=black, very thick] coordinates {(P,694.618)};

%Salinas
\nextgroupplot[
ylabel={Time (s)},
ybar=-5pt,
bar width=0.2cm,
xtick=data,
]
\addplot[orange, fill,draw=black, very thick] coordinates {(A,6.855)};
\addplot[blue, fill,draw=black, very thick] coordinates {(B,13.373)};
\addplot[teal, fill,draw=black, very thick] coordinates {(C,110.908)};
\addplot[lime, fill,draw=black, very thick] coordinates {(D,1024.356)};
\addplot[brown, fill,draw=black, very thick]  coordinates {(E,10.628)};
\addplot[gray1, fill,draw=black, very thick]  coordinates {(F,7.654)};
\addplot[gray2, fill,draw=black, very thick] coordinates {(G,8.065)};
\addplot[gray3, fill,draw=black, very thick] coordinates {(H,164.868)};
\addplot[gray4, fill,draw=black, very thick] coordinates {(I,10.512)};
\addplot[gray5, fill,draw=black, very thick] coordinates {(J,165.885)};
\addplot[gray6, fill,draw=black, very thick] coordinates {(K,163.114)};
\addplot[hotmagenta, fill,draw=black, very thick] coordinates {(L,159.950)};
\addplot[yellow, fill,draw=black, very thick] coordinates {(M,163.529)};
\addplot[red, fill,draw=black, very thick] coordinates {(N,155.838)};
\addplot[cyan, fill,draw=black, very thick] coordinates {(O,214.298)};
\addplot[green, fill,draw=black, very thick] coordinates {(P,306.147)};

\nextgroupplot[
ybar=-5pt,
bar width=0.2cm,
xtick=data,
]
\addplot[orange, fill,draw=black, very thick] coordinates {(A,7.141)};
\addplot[blue, fill,draw=black, very thick] coordinates {(B,1.181)};
\addplot[teal, fill,draw=black, very thick] coordinates {(C,17.889)};
\addplot[lime, fill,draw=black, very thick] coordinates {(D,167.373)};
\addplot[brown, fill,draw=black, very thick]  coordinates {(E,3.600)};
\addplot[gray1, fill,draw=black, very thick]  coordinates {(F,1.895)};
\addplot[gray2, fill,draw=black, very thick] coordinates {(G,2.022)};
\addplot[gray3, fill,draw=black, very thick] coordinates {(H,25.499)};
\addplot[gray4, fill,draw=black, very thick] coordinates {(I,2.893)};
\addplot[gray5, fill,draw=black, very thick] coordinates {(J,26.298)};
\addplot[gray6, fill,draw=black, very thick] coordinates {(K,24.695)};
\addplot[hotmagenta, fill,draw=black, very thick] coordinates {(L,27.794)};
\addplot[yellow, fill,draw=black, very thick] coordinates {(M,25.234)};
\addplot[red, fill,draw=black, very thick] coordinates {(N,24.348)};
\addplot[cyan, fill,draw=black, very thick] coordinates {(O,32.997)};
\addplot[green, fill,draw=black, very thick] coordinates {(P,49.881)};

\end{groupplot}
\node[anchor=south, above=0.0cm] at ($(my plots c1r1.north)$){\tiny 100\%};\node[anchor=south, above=0.0cm] at ($(my plots c2r1.north)$){\tiny 66\%};\node[anchor=south, above=0.0cm] at ($(my plots c1r2.north)$){\tiny 33\%};\node[anchor=south, above=0.0cm] at ($(my plots c2r2.north)$){\tiny 13\%};\node[anchor=south, above=0.0cm] at ($(my plots c1r3.north)$){\tiny 6\%};\node[anchor=south, above=0.0cm] at ($(my plots c2r3.north)$){\tiny 1\%};
\end{tikzpicture}
\caption{Average train. time for Urban (all $\TrainingSet$'s sizes):
\legendsquare{fill=orange}\,LMM\,\cite{heinz1999fully},
\legendsquare{fill=blue}\,SVR\,\cite{6638039}
\legendsquare{fill=teal}\,CB-CNN\,\cite{cnn_unmixing},
\legendsquare{fill=lime}\,WS-AE\,\cite{weaklyae},
\legendsquare{fill=brown}\,UnDIP\,\cite{Rasti2021undip},
\legendsquare{fill=gray1}\,MB(1D),
\legendsquare{fill=gray2}\,MB(2D),
\legendsquare{fill=gray3}\,MB(3D),
\legendsquare{fill=gray4}\,MB(1D+2D),
\legendsquare{fill=gray5}\,MB(1D+3D),
\legendsquare{fill=gray6}\,MB(2D+3D),
\legendsquare{fill=hotmagenta}\,MB,
\legendsquare{fill=yellow}\,MB-DR,
\legendsquare{fill=red}\,MB-Res,
\legendsquare{fill=cyan}\,MB-TL,
\legendsquare{fill=green}\,MB-PT.}
\label{fig:times}
\end{figure}
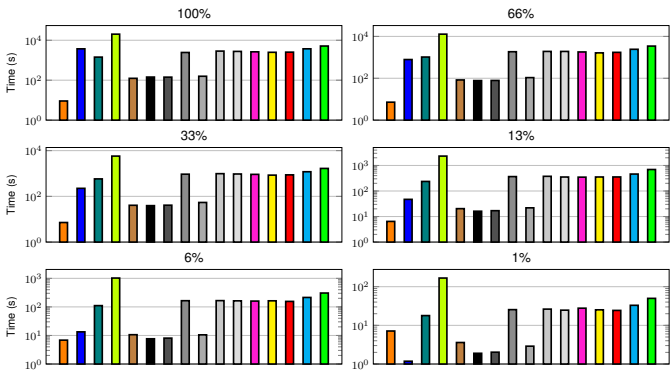

In Fig.~\ref{fig:times}, we can appreciate that the training time of MB CNN and its variants, averaged across all independent executions for Urban, being the largest set, is shorter or comparable to other techniques. Similarly, the averaged inference time (across all training set sizes) over all test samples in the test sets amounted to 3.6--4.1~s for MB (it was 0.2--3.8~s for the variants utilizing a subset of all branches)---it is up to $57\times$ faster than SVR (the average inference time for LMM, CB-CNN, and WS-AE was 12.3~s, 2.1~s, and 24.9~s). It shows that the multi-branch processing can be executed in short time.

%%%%%%%%%%%%%%%%%%%%%%%%%%

\section{Conclusions and Future Work}\label{sec:conclusions}

Hyperspectral unmixing remains one of the most challenging tasks in HSI analysis. Although deep learning has been blooming in the field, new deep architectures emerge at a steady pace to improve the quality of HU. We introduced a multi-branch CNN for HU that benefits from automated representation learning and efficient fusion of spectral, spatial, and spectral-spatial features during the unmixing process. The experiments showed that our technique outperforms other classical and deep learning methods, and indicated the benefits of utilizing parallel branches that capture spectral, spatial and spectral-spatial features which are later fused in the network.

Although our experiments involved the extensive multi-fold analysis, such training/test samples are drawn from the very same HSI (i.e.,~they capture the same Earth peculiarities). Thus, acquiring spatially de-correlated HSIs for HU validation would help us better understand the generalization of such methods. It would be interesting to enhance the multi-branch models in order to perform the endmember determination, as it is a pivotal step in real-life Earth observation use cases, and to deploy them for the non-linear models, e.g., in microscopic scenarios~\cite{Rasti2021undip}. Finally, to reduce the amount of data to be transferred from a satellite equipped with a hyperspectral imager, and to accelerate the response time through the in-orbit analysis of raw HSIs, we are working on deploying the MB CNNs onboard KP Labs' Intuition-1 hyperspectral mission.

\ifCLASSOPTIONcaptionsoff
  \newpage
\fi

\bibliographystyle{IEEEtran}
\bibliography{ref_all}

\end{document}

\begin{center} \textbf{A Multibranch Convolutional Neural Network for Hyperspectral Unmixing (Supplementary Material)}\end{center}
\begin{center} Lukasz Tulczyjew, Michal Kawulok, Nicolas Long{\'e}p{\'e}, Bertrand Le Saux, Jakub Nalepa\end{center}
\begin{center} \texttt{jnalepa@ieee.org}\end{center}

This supplementary material collects the detailed experimental results obtained using the investigated hyperspectral unmixing techniques (Section~\ref{sec:exp}). To access the diagrams for each of the implemented networks, the code to reproduce the experiments, and other detailed results of the experimental study (Wilcoxon tests, training and test times for each training set size and benchmark set, and the detailed metric values), see \url{https://gitlab.com/jnalepa/mbhu}.

\section{Detailed Experimental Results}\label{sec:exp}

\begin{figure}[ht!]
    \centering
    \includegraphics[width=0.8\columnwidth]{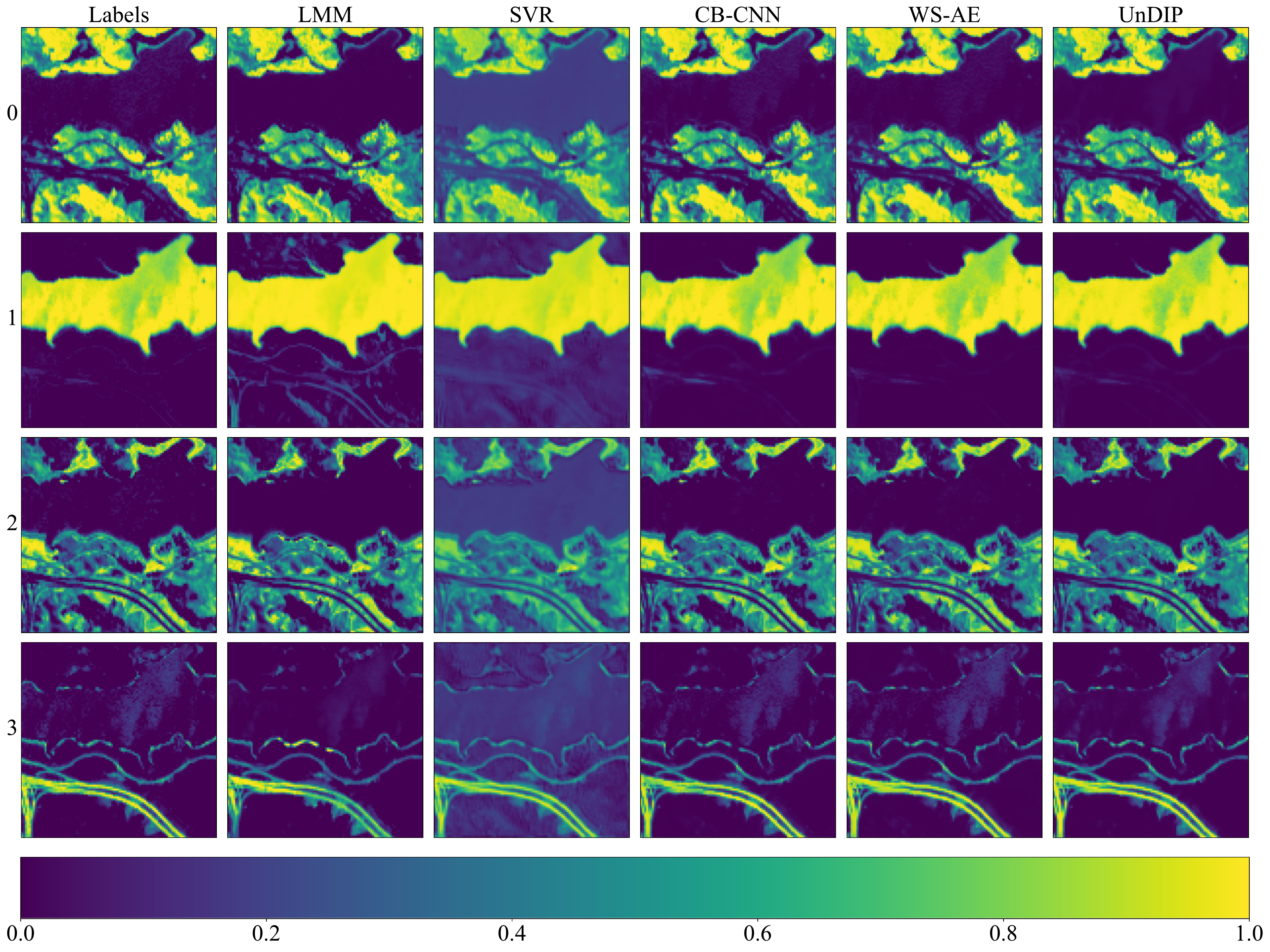}
    \caption{The abundance maps obtained by applying different unmixing techniques from the literature (trained over the entire training set) over Jasper Ridge (the rows correspond to the endmembers: 0---tree, 1---water, 2---soil, 3---road). The \textit{Labels} column corresponds to the ground truth.}
    \label{fig:pred_literature}
\end{figure}

\begin{figure}[ht!]
    \centering
    \includegraphics[width=0.8\columnwidth]{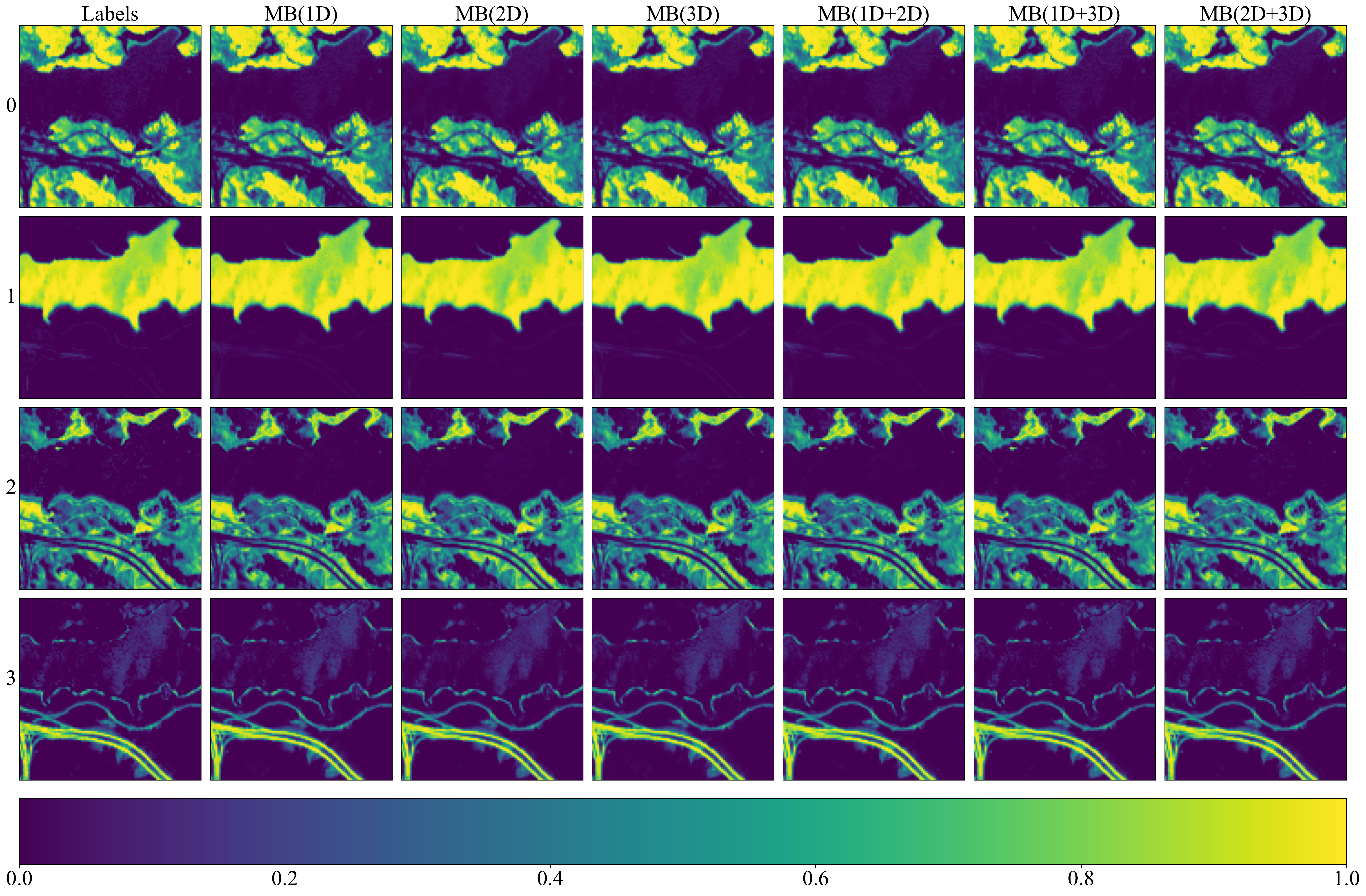}
    \caption{The abundance maps obtained by applying different variants of the proposed multi-branch architecture (investigated in the ablation study, and trained over the entire training set) over Jasper Ridge (the rows correspond to the endmembers: 0---tree, 1---water, 2---soil, 3---road). The \textit{Labels} column corresponds to the ground truth.}
    \label{fig:pred_variants}
\end{figure}

\begin{figure}[ht!]
    \centering
    \includegraphics[width=0.8\columnwidth]{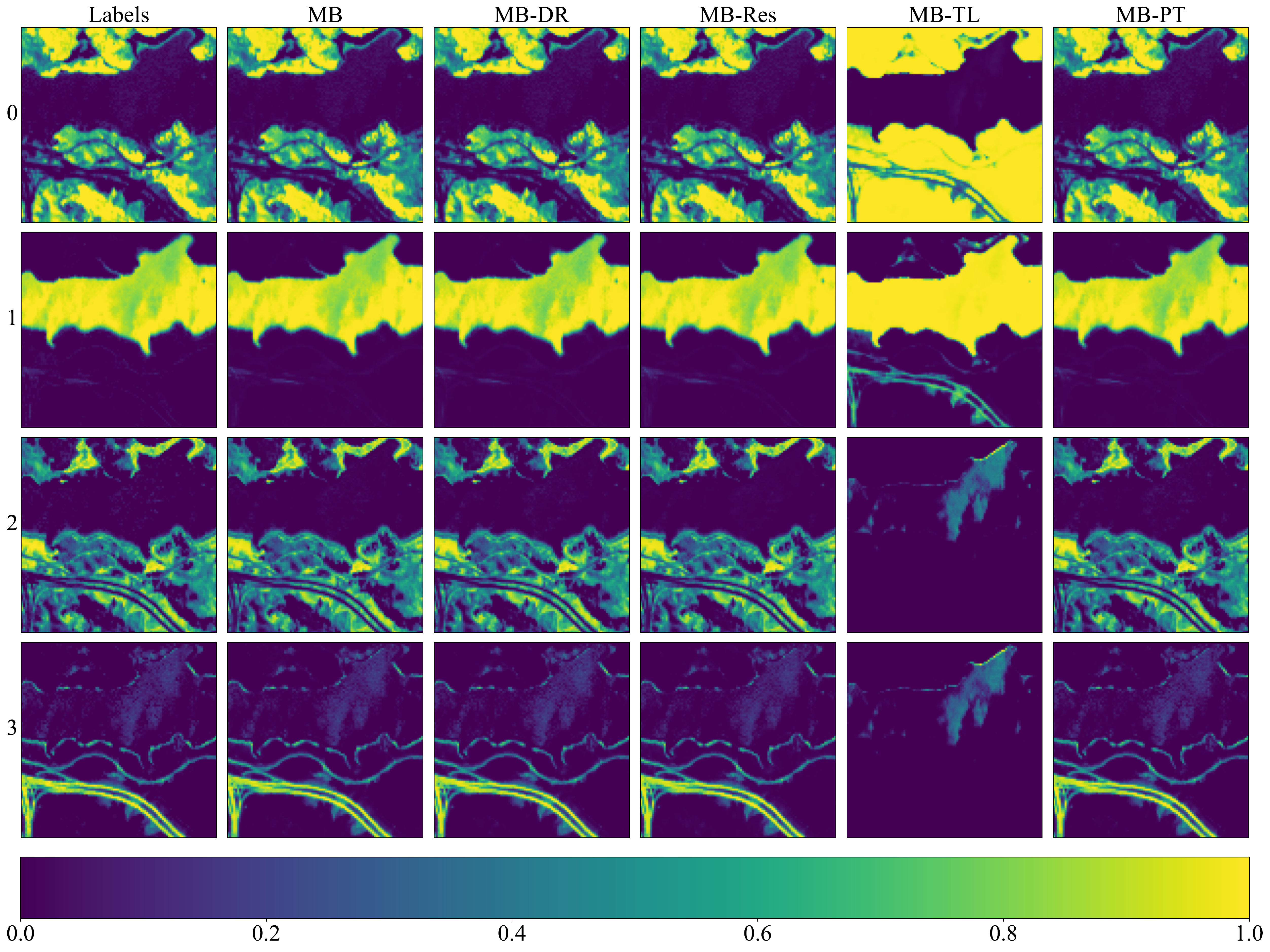}
    \caption{The abundance maps obtained by different variants of our multi-branch architecture (trained over the entire training set) over Jasper Ridge (the rows correspond to the endmembers: 0---tree, 1---water, 2---soil, 3---road). The \textit{Labels} column corresponds to the ground truth.}
    \label{fig:pred_ours}
\end{figure}

\begin{figure}[ht!]
    \centering
    \includegraphics[width=0.65\columnwidth]{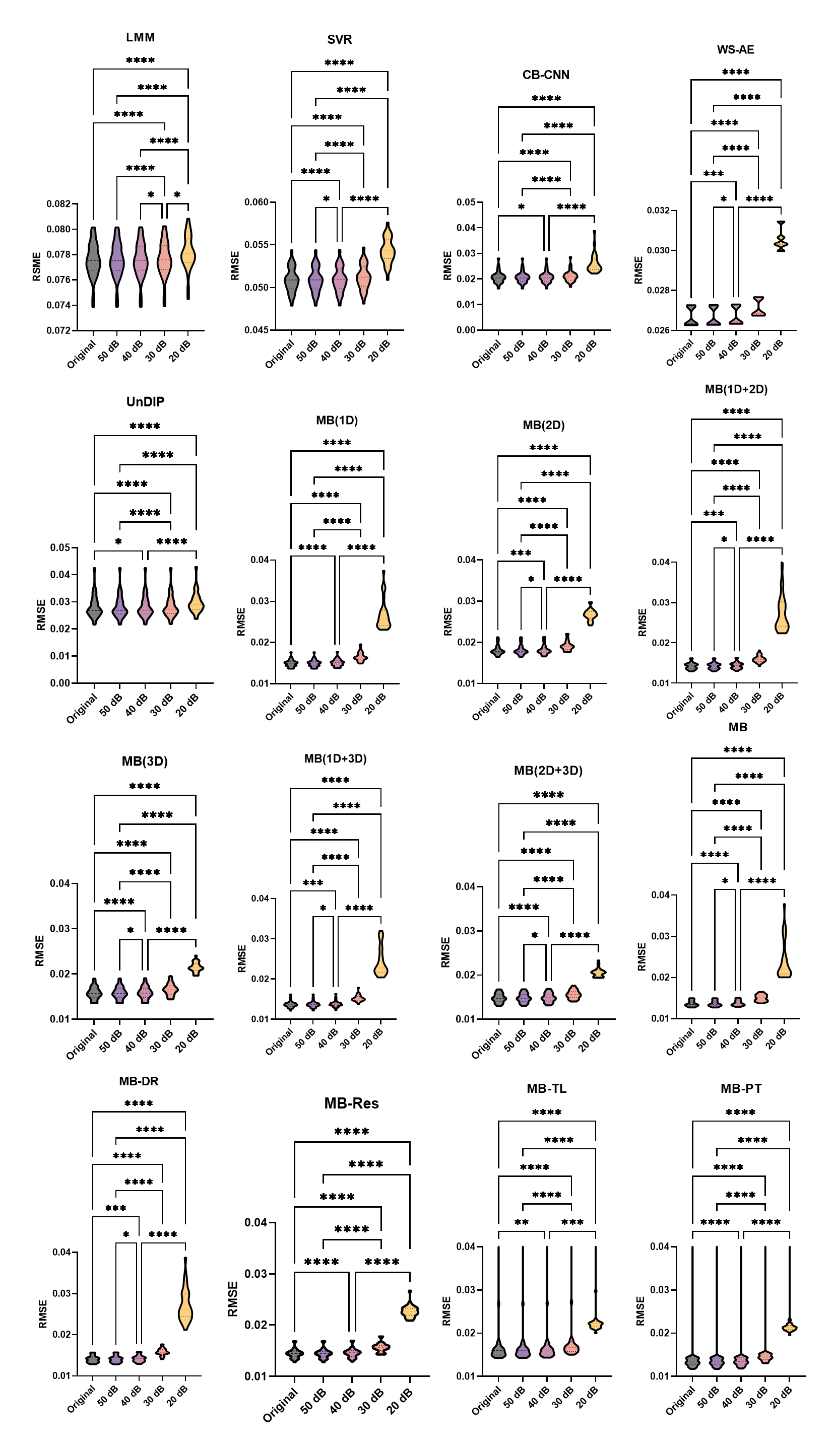}
    \caption{The impact of the white zero-mean Gaussian noise added to the original data (with the signal-to-noise ratio of 20, 30, 40, and 50 dB) on the performance (quantified as RMSE) of all investigated unmixing algorithms over the Jasper Ridge dataset. The results of the Friedman's tests with post-hoc Dunn's, verifying if the differences are statistically significant, are reported as: * ($p<0.05$), ** ($p<0.01$), *** ($p<0.001$), and **** ($p<0.0001$). }
    \label{fig:noise}
\end{figure}

In this section, we gather the detailed experimental results that are discussed in the main body of the letter. The following results are included in the supplementary material:
\begin{itemize}
    \item Overall RMSE over Samson reported for different training sizes and all investigated algorithms: Figure~\ref{fig:rmse_samson}.
    \item Overall rmsAAD over the Samson, Urban, and Jasper Ridge datasets reported for different training sizes and all investigated algorithms: Figures~\ref{fig:rmsAAD_samson}--\ref{fig:rmsaad_jasper}.
    \item The ranking (RMSE and rmsAAD) obtained for the Samson, Urban, and Jasper Ridge datasets by all investigated HU algorithms: Tables~\ref{tab:ranking_samson}--\ref{tab:ranking_jasper}.
    \item The abundance maps obtained by applying different unmixing techniques from the literature, and by different variants of the proposed multi-branch architecture (trained over the entire training set) over the Jasper Ridge dataset: Figures~\ref{fig:pred_literature}--\ref{fig:pred_ours}.
    \item The impact of the white zero-mean Gaussian noise added to the original data (with the signal-to-noise ratio of 20, 30, 40, and 50 dB) on the performance (quantified as RMSE) of all investigated unmixing algorithms over the Jasper Ridge dataset: Figure~\ref{fig:noise}.
    \item The impact of the white zero-mean Gaussian noise added to the original data (with the signal-to-noise ratio of 20, 30, 40, and 50 dB) on the performance (quantified as mean, median, and standard deviation of RMSE) of all investigated unmixing algorithms over the Jasper Ridge dataset: Table~\ref{tab:noise_impact}.
    \item The impact of the patch size on RMSE (on Jasper Ridge), quantified as $\Delta_{\rm RMSE}={\rm RMSE}^{k\times k}-{\rm RMSE}^{3\times 3}$ ($k=\{5, 7, 9\}$) obtained using MB-DR, MB-Res, MB-TL, and MB-PT, trained from the training sets of various sizes: Tables~\ref{tab:patch_size_mbdr}--\ref{tab:patch_size_mbpt}.
\end{itemize}

\begin{figure}[ht!]
\pgfplotsset{every axis title/.append style={at={(0.5,0.9)}}}
\center
\begin{tikzpicture}[scale=0.8]		
[font=\footnotesize\sffamily]
\begin{groupplot}[
group style={group size=2 by 3, horizontal sep=0.8cm, vertical sep=0.45cm, group name=my plots},
legend style={at={(0.5,-0.15)},
anchor=north,legend columns=-1},
symbolic x coords={A,B,C,D,E,F,G,H,I,J,K,L,M,N,O,P},
x=0.42cm,xticklabels=\empty,
height=4cm,xtick style={draw=none},
ymajorgrids,y tick label style={/pgf/number format/.cd,fixed,fixed zerofill,precision=2,/tikz/.cd}]
\nextgroupplot[
ylabel={Overall RMSE},
ybar=-5pt,
bar width=0.2cm,
xtick=data,ymax=0.06000,ymin=0,title=100\%
]
\addplot[orange, fill,draw=black, very thick] coordinates {(A,      0.053   )};
\draw node[above,xshift=-2mm,yshift=20.5mm] {\scriptsize$\uparrow$0.42};
\addplot[blue, fill,draw=black, very thick] coordinates {(B,        0.053   )};
\addplot[teal, fill,draw=black, very thick] coordinates {(C,        0.019   )};
\addplot[lime, fill,draw=black, very thick] coordinates {(D,        0.022   )};
\addplot[brown, fill,draw=black, very thick]  coordinates {(E,      0.036   )};
\addplot[gray1, fill,draw=black, very thick]  coordinates {(F,      0.012   )};
\addplot[gray2, fill,draw=black, very thick] coordinates {(G,       0.015   )};
\addplot[gray3, fill,draw=black, very thick] coordinates {(H,       0.013   )};
\addplot[gray4, fill,draw=black, very thick] coordinates {(I,       0.010   )};
\addplot[gray5, fill,draw=black, very thick] coordinates {(J,       0.010   )};
\addplot[gray6, fill,draw=black, very thick] coordinates {(K,       0.011   )};
\addplot[hotmagenta, fill,draw=black, very thick] coordinates {(L,  0.010   )};
\addplot[yellow, fill,draw=black, very thick] coordinates {(M,      0.010   )};
\addplot[red, fill,draw=black, very thick] coordinates {(N,         0.010   )};
\addplot[cyan, fill,draw=black, very thick] coordinates {(O,        0.014   )};
\addplot[green, fill,draw=black, very thick] coordinates {(P,       0.009   )};

\nextgroupplot[
ybar=-5pt,
bar width=0.2cm,
xtick=data,ymax=0.062,ymin=0,title=66\%
]
\draw node[above,xshift=-2mm,yshift=20.5mm] {\scriptsize$\uparrow$0.42};
\addplot[orange, fill,draw=black, very thick] coordinates {(A,      0.055   )};
\addplot[blue, fill,draw=black, very thick] coordinates {(B,        0.055   )};
\addplot[teal, fill,draw=black, very thick] coordinates {(C,        0.020   )};
\addplot[lime, fill,draw=black, very thick] coordinates {(D,        0.023   )};
\addplot[brown, fill,draw=black, very thick]  coordinates {(E,      0.045   )};
\addplot[gray1, fill,draw=black, very thick]  coordinates {(F,      0.014   )};
\addplot[gray2, fill,draw=black, very thick] coordinates {(G,       0.018   )};
\addplot[gray3, fill,draw=black, very thick] coordinates {(H,       0.014   )};
\addplot[gray4, fill,draw=black, very thick] coordinates {(I,       0.012   )};
\addplot[gray5, fill,draw=black, very thick] coordinates {(J,       0.013   )};
\addplot[gray6, fill,draw=black, very thick] coordinates {(K,       0.012   )};
\addplot[hotmagenta, fill,draw=black, very thick] coordinates {(L,  0.012   )};
\addplot[yellow, fill,draw=black, very thick] coordinates {(M,      0.012   )};
\addplot[red, fill,draw=black, very thick] coordinates {(N,         0.012   )};
\addplot[cyan, fill,draw=black, very thick] coordinates {(O,        0.028   )};
\addplot[green, fill,draw=black, very thick] coordinates {(P,       0.025   )};

\nextgroupplot[
ylabel={RMSE},
ybar=-5pt,
bar width=0.2cm,
xtick=data,ymax=0.11,ymin=0,title=33\%
]
\draw node[above,xshift=-2mm,yshift=20.5mm] {\scriptsize$\uparrow$0.42};
\addplot[orange, fill,draw=black, very thick] coordinates {(A,      0.096   )};
\addplot[blue, fill,draw=black, very thick] coordinates {(B,        0.060   )};
\addplot[teal, fill,draw=black, very thick] coordinates {(C,        0.023   )};
\addplot[lime, fill,draw=black, very thick] coordinates {(D,        0.027   )};
\addplot[brown, fill,draw=black, very thick]  coordinates {(E,      0.080   )};
\addplot[gray1, fill,draw=black, very thick]  coordinates {(F,      0.018   )};
\addplot[gray2, fill,draw=black, very thick] coordinates {(G,       0.026   )};
\addplot[gray3, fill,draw=black, very thick] coordinates {(H,       0.019   )};
\addplot[gray4, fill,draw=black, very thick] coordinates {(I,       0.017   )};
\addplot[gray5, fill,draw=black, very thick] coordinates {(J,       0.017   )};
\addplot[gray6, fill,draw=black, very thick] coordinates {(K,       0.018   )};
\addplot[hotmagenta, fill,draw=black, very thick] coordinates {(L,  0.016   )};
\addplot[yellow, fill,draw=black, very thick] coordinates {(M,      0.017   )};
\addplot[red, fill,draw=black, very thick] coordinates {(N,         0.017   )};
\addplot[cyan, fill,draw=black, very thick] coordinates {(O,        0.096   )};
\addplot[green, fill,draw=black, very thick] coordinates {(P,       0.017   )};

\nextgroupplot[
ybar=-5pt,
bar width=0.2cm,
xtick=data,ymax=0.18,ymin=0,title=13\%
]
%\addplot[black,line legend,sharp plot,nodes near coords={},
%    update limits=false,shorten >=-3mm,shorten <=-3mm] 
%    coordinates {(A,0.024) (P,0.024)} 
%    node[midway,below,font=\bfseries\sffamily]{min};
\draw node[above,xshift=-2mm,yshift=20.5mm] {\scriptsize$\uparrow$0.42};
\addplot[orange, fill,draw=black, very thick] coordinates {(A,      0.160   )};
\addplot[blue, fill,draw=black, very thick] coordinates {(B,        0.069   )};
\addplot[teal, fill,draw=black, very thick] coordinates {(C,        0.032   )};
\addplot[lime, fill,draw=black, very thick] coordinates {(D,        0.036   )};
\addplot[brown, fill,draw=black, very thick]  coordinates {(E,      0.160   )};
\addplot[gray1, fill,draw=black, very thick]  coordinates {(F,      0.026   )};
\addplot[gray2, fill,draw=black, very thick] coordinates {(G,       0.037   )};
\addplot[gray3, fill,draw=black, very thick] coordinates {(H,       0.028   )};
\addplot[gray4, fill,draw=black, very thick] coordinates {(I,       0.025   )};
\addplot[gray5, fill,draw=black, very thick] coordinates {(J,       0.024   )};
\addplot[gray6, fill,draw=black, very thick] coordinates {(K,       0.026   )};
\addplot[hotmagenta, fill,draw=black, very thick] coordinates {(L,  0.024   )};
\addplot[yellow, fill,draw=black, very thick] coordinates {(M,      0.025   )};
\addplot[red, fill,draw=black, very thick] coordinates {(N,         0.025   )};
\addplot[cyan, fill,draw=black, very thick] coordinates {(O,        0.152   )};
\addplot[green, fill,draw=black, very thick] coordinates {(P,       0.038   )};

\nextgroupplot[
ylabel={RMSE},
ybar=-5pt,
bar width=0.2cm,
xtick=data,ymax=0.21,title=6\%
]
\draw node[above,xshift=-2mm,yshift=18mm] {\scriptsize$\uparrow$0.42};
\draw node[above,xshift=+60mm,yshift=18mm] {\scriptsize$\uparrow$0.24};
\addplot[orange, fill,draw=black, very thick] coordinates {(A,      0.183   )};
\addplot[blue, fill,draw=black, very thick] coordinates {(B,        0.079   )};
\addplot[teal, fill,draw=black, very thick] coordinates {(C,        0.043   )};
\addplot[lime, fill,draw=black, very thick] coordinates {(D,        0.043   )};
\addplot[brown, fill,draw=black, very thick]  coordinates {(E,      0.183   )};
\addplot[gray1, fill,draw=black, very thick]  coordinates {(F,      0.037   )};
\addplot[gray2, fill,draw=black, very thick] coordinates {(G,       0.044   )};
\addplot[gray3, fill,draw=black, very thick] coordinates {(H,       0.039   )};
\addplot[gray4, fill,draw=black, very thick] coordinates {(I,       0.037   )};
\addplot[gray5, fill,draw=black, very thick] coordinates {(J,       0.034   )};
\addplot[gray6, fill,draw=black, very thick] coordinates {(K,       0.038   )};
\addplot[hotmagenta, fill,draw=black, very thick] coordinates {(L,  0.034   )};
\addplot[yellow, fill,draw=black, very thick] coordinates {(M,      0.035   )};
\addplot[red, fill,draw=black, very thick] coordinates {(N,         0.035   )};
\addplot[cyan, fill,draw=black, very thick] coordinates {(O,        0.183   )};
\addplot[green, fill,draw=black, very thick] coordinates {(P,       0.049   )};

\nextgroupplot[
ybar=-5pt,
bar width=0.2cm,
xtick=data,ymax=0.35,ymin=0,title=1\%
]
\draw node[above,xshift=-2mm,yshift=20mm] {\scriptsize$\uparrow$0.42};
\addplot[orange, fill,draw=black, very thick] coordinates {(A,      0.302   )};
\addplot[blue, fill,draw=black, very thick] coordinates {(B,        0.113   )};
\addplot[teal, fill,draw=black, very thick] coordinates {(C,        0.075   )};
\addplot[lime, fill,draw=black, very thick] coordinates {(D,        0.125   )};
\addplot[brown, fill,draw=black, very thick]  coordinates {(E,      0.302   )};
\addplot[gray1, fill,draw=black, very thick]  coordinates {(F,      0.104   )};
\addplot[gray2, fill,draw=black, very thick] coordinates {(G,       0.116   )};
\addplot[gray3, fill,draw=black, very thick] coordinates {(H,       0.091   )};
\addplot[gray4, fill,draw=black, very thick] coordinates {(I,       0.099   )};
\addplot[gray5, fill,draw=black, very thick] coordinates {(J,       0.078   )};
\addplot[gray6, fill,draw=black, very thick] coordinates {(K,       0.102   )};
\addplot[hotmagenta, fill,draw=black, very thick] coordinates {(L,  0.071   )};
\addplot[yellow, fill,draw=black, very thick] coordinates {(M,      0.086   )};
\addplot[red, fill,draw=black, very thick] coordinates {(N,         0.084   )};
\addplot[cyan, fill,draw=black, very thick] coordinates {(O,        0.311   )};
\addplot[green, fill,draw=black, very thick] coordinates {(P,       0.165   )};
\end{groupplot}
\end{tikzpicture}
\caption{Overall RMSE over Samson reported for different training sizes:
\legendsquare{fill=orange}~LMM,
\legendsquare{fill=blue}~SVR,
\legendsquare{fill=teal}~CB-CNN,
\legendsquare{fill=lime}~WS-AE,
\legendsquare{fill=brown}~UnDIP,
\legendsquare{fill=gray1}~MB(1D),
\legendsquare{fill=gray2}~MB(2D),
\legendsquare{fill=gray3}~MB(3D),
\legendsquare{fill=gray4}~MB(1D+2D),
\legendsquare{fill=gray5}~MB(1D+3D),
\legendsquare{fill=gray6}~MB(2D+3D),
\legendsquare{fill=hotmagenta}~MB,
\legendsquare{fill=yellow}~MB-DR,
\legendsquare{fill=red}~MB-Res,
\legendsquare{fill=cyan}~MB-TL,
\legendsquare{fill=green}~MB-PT. For some methods, we indicate the exact value of RMSE above the arrow to maintain readability of the plot (those values are outside the current RMSE range on the Y axis).}
\label{fig:rmse_samson}
\end{figure}

\begin{figure}[ht!]
\pgfplotsset{every axis title/.append style={at={(0.5,0.9)}}}
\center
\begin{tikzpicture}[scale=0.8]		
[font=\footnotesize\sffamily]
\begin{groupplot}[
group style={group size=2 by 3, horizontal sep=0.8cm, vertical sep=0.45cm, group name=my plots},
legend style={at={(0.5,-0.15)},
anchor=north,legend columns=-1},
symbolic x coords={A,B,C,D,E,F,G,H,I,J,K,L,M,N,O,P},
x=0.42cm,xticklabels=\empty,
height=4cm,xtick style={draw=none},
ymajorgrids,y tick label style={/pgf/number format/.cd,fixed,fixed zerofill,precision=2,/tikz/.cd}]
\nextgroupplot[
ylabel={rmsAAD},
ybar=-5pt,
bar width=0.2cm,
xtick=data,ymax=0.12,ymin=0,title=100\%
]
\draw node[above,xshift=-2mm,yshift=20mm] {\scriptsize$\uparrow$0.95};
\addplot[orange, fill,draw=black, very thick] coordinates {(A,      0.102   )};
\addplot[blue, fill,draw=black, very thick] coordinates {(B,        0.102   )};
\addplot[teal, fill,draw=black, very thick] coordinates {(C,        0.037   )};
\addplot[lime, fill,draw=black, very thick] coordinates {(D,        0.043   )};
\addplot[brown, fill,draw=black, very thick]  coordinates {(E,      0.073   )};
\addplot[gray1, fill,draw=black, very thick]  coordinates {(F,      0.023   )};
\addplot[gray2, fill,draw=black, very thick] coordinates {(G,       0.031   )};
\addplot[gray3, fill,draw=black, very thick] coordinates {(H,       0.025   )};
\addplot[gray4, fill,draw=black, very thick] coordinates {(I,       0.020   )};
\addplot[gray5, fill,draw=black, very thick] coordinates {(J,       0.020   )};
\addplot[gray6, fill,draw=black, very thick] coordinates {(K,       0.023   )};
\addplot[hotmagenta, fill,draw=black, very thick] coordinates {(L,  0.019   )};
\addplot[yellow, fill,draw=black, very thick] coordinates {(M,      0.020   )};
\addplot[red, fill,draw=black, very thick] coordinates {(N,         0.019   )};
\addplot[cyan, fill,draw=black, very thick] coordinates {(O,        0.027   )};
\addplot[green, fill,draw=black, very thick] coordinates {(P,       0.017   )};

\nextgroupplot[
ybar=-5pt,
bar width=0.2cm,
xtick=data,ymax=0.12,ymin=0,title=66\%
]
\draw node[above,xshift=-2mm,yshift=20.5mm] {\scriptsize$\uparrow$0.95};
\addplot[orange, fill,draw=black, very thick] coordinates {(A,      0.107   )};
\addplot[blue, fill,draw=black, very thick] coordinates {(B,        0.107   )};
\addplot[teal, fill,draw=black, very thick] coordinates {(C,        0.039   )};
\addplot[lime, fill,draw=black, very thick] coordinates {(D,        0.044   )};
\addplot[brown, fill,draw=black, very thick]  coordinates {(E,      0.088   )};
\addplot[gray1, fill,draw=black, very thick]  coordinates {(F,      0.027   )};
\addplot[gray2, fill,draw=black, very thick] coordinates {(G,       0.036   )};
\addplot[gray3, fill,draw=black, very thick] coordinates {(H,       0.029   )};
\addplot[gray4, fill,draw=black, very thick] coordinates {(I,       0.025   )};
\addplot[gray5, fill,draw=black, very thick] coordinates {(J,       0.025   )};
\addplot[gray6, fill,draw=black, very thick] coordinates {(K,       0.026   )};
\addplot[hotmagenta, fill,draw=black, very thick] coordinates {(L,  0.023   )};
\addplot[yellow, fill,draw=black, very thick] coordinates {(M,      0.024   )};
\addplot[red, fill,draw=black, very thick] coordinates {(N,         0.025   )};
\addplot[cyan, fill,draw=black, very thick] coordinates {(O,        0.059   )};
\addplot[green, fill,draw=black, very thick] coordinates {(P,       0.049   )};

\nextgroupplot[
ylabel={rmsAAD},
ybar=-5pt,
bar width=0.2cm,
xtick=data,ymax=0.135,ymin=0,title=33\%
]
\draw node[above,xshift=-2mm,yshift=20.5mm] {\scriptsize$\uparrow$0.95};
\draw node[above,xshift=16mm,yshift=20.5mm] {\scriptsize$\uparrow$0.16};
\draw node[above,xshift=60mm,yshift=20.5mm] {\scriptsize$\uparrow$0.21};
\addplot[orange, fill,draw=black, very thick] coordinates {(A,      0.118   )};
\addplot[blue, fill,draw=black, very thick] coordinates {(B,        0.118   )};
\addplot[teal, fill,draw=black, very thick] coordinates {(C,        0.047   )};
\addplot[lime, fill,draw=black, very thick] coordinates {(D,        0.051   )};
\addplot[brown, fill,draw=black, very thick]  coordinates {(E,      0.118   )};
\addplot[gray1, fill,draw=black, very thick]  coordinates {(F,      0.036   )};
\addplot[gray2, fill,draw=black, very thick] coordinates {(G,       0.055   )};
\addplot[gray3, fill,draw=black, very thick] coordinates {(H,       0.037   )};
\addplot[gray4, fill,draw=black, very thick] coordinates {(I,       0.034   )};
\addplot[gray5, fill,draw=black, very thick] coordinates {(J,       0.034   )};
\addplot[gray6, fill,draw=black, very thick] coordinates {(K,       0.036   )};
\addplot[hotmagenta, fill,draw=black, very thick] coordinates {(L,  0.032   )};
\addplot[yellow, fill,draw=black, very thick] coordinates {(M,      0.034   )};
\addplot[red, fill,draw=black, very thick] coordinates {(N,         0.034   )};
\addplot[cyan, fill,draw=black, very thick] coordinates {(O,        0.118   )};
\addplot[green, fill,draw=black, very thick] coordinates {(P,       0.033   )};

\nextgroupplot[
ybar=-5pt,
bar width=0.2cm,
xtick=data,ymax=0.16,ymin=0,title=13\%
]
%\addplot[black,line legend,sharp plot,nodes near coords={},
%    update limits=false,shorten >=-3mm,shorten <=-3mm] 
%    coordinates {(A,0.024) (P,0.024)} 
%    node[midway,below,font=\bfseries\sffamily]{min};
\draw node[above,xshift=-2mm,yshift=20.5mm] {\scriptsize$\uparrow$0.95};
\draw node[above,xshift=16mm,yshift=20.5mm] {\scriptsize$\uparrow$0.31};
\draw node[above,xshift=60mm,yshift=20.5mm] {\scriptsize$\uparrow$0.32};
\addplot[orange, fill,draw=black, very thick] coordinates {(A,      0.138   )};
\addplot[blue, fill,draw=black, very thick] coordinates {(B,        0.138   )};
\addplot[teal, fill,draw=black, very thick] coordinates {(C,        0.068   )};
\addplot[lime, fill,draw=black, very thick] coordinates {(D,        0.073   )};
\addplot[brown, fill,draw=black, very thick]  coordinates {(E,      0.138   )};
\addplot[gray1, fill,draw=black, very thick]  coordinates {(F,      0.054   )};
\addplot[gray2, fill,draw=black, very thick] coordinates {(G,       0.081   )};
\addplot[gray3, fill,draw=black, very thick] coordinates {(H,       0.057   )};
\addplot[gray4, fill,draw=black, very thick] coordinates {(I,       0.053   )};
\addplot[gray5, fill,draw=black, very thick] coordinates {(J,       0.049   )};
\addplot[gray6, fill,draw=black, very thick] coordinates {(K,       0.055   )};
\addplot[hotmagenta, fill,draw=black, very thick] coordinates {(L,  0.049   )};
\addplot[yellow, fill,draw=black, very thick] coordinates {(M,      0.051   )};
\addplot[red, fill,draw=black, very thick] coordinates {(N,         0.051   )};
\addplot[cyan, fill,draw=black, very thick] coordinates {(O,        0.138   )};
\addplot[green, fill,draw=black, very thick] coordinates {(P,       0.078   )};

\nextgroupplot[
ylabel={rmsAAD},
ybar=-5pt,
bar width=0.2cm,
xtick=data,ymax=0.175,title=6\%
]
\draw node[above,xshift=-2mm,yshift=18mm] {\scriptsize$\uparrow$0.95};
\draw node[above,xshift=16mm,yshift=18mm] {\scriptsize$\uparrow$0.38};
\draw node[above,xshift=60mm,yshift=18mm] {\scriptsize$\uparrow$0.49};
\addplot[orange, fill,draw=black, very thick] coordinates {(A,      0.159   )};
\addplot[blue, fill,draw=black, very thick] coordinates {(B,        0.159   )};
\addplot[teal, fill,draw=black, very thick] coordinates {(C,        0.090   )};
\addplot[lime, fill,draw=black, very thick] coordinates {(D,        0.091   )};
\addplot[brown, fill,draw=black, very thick]  coordinates {(E,      0.159   )};
\addplot[gray1, fill,draw=black, very thick]  coordinates {(F,      0.079   )};
\addplot[gray2, fill,draw=black, very thick] coordinates {(G,       0.095   )};
\addplot[gray3, fill,draw=black, very thick] coordinates {(H,       0.082   )};
\addplot[gray4, fill,draw=black, very thick] coordinates {(I,       0.080   )};
\addplot[gray5, fill,draw=black, very thick] coordinates {(J,       0.073   )};
\addplot[gray6, fill,draw=black, very thick] coordinates {(K,       0.081   )};
\addplot[hotmagenta, fill,draw=black, very thick] coordinates {(L,  0.072   )};
\addplot[yellow, fill,draw=black, very thick] coordinates {(M,      0.074   )};
\addplot[red, fill,draw=black, very thick] coordinates {(N,         0.073   )};
\addplot[cyan, fill,draw=black, very thick] coordinates {(O,        0.159   )};
\addplot[green, fill,draw=black, very thick] coordinates {(P,       0.102   )};

\nextgroupplot[
ybar=-5pt,
bar width=0.2cm,
xtick=data,ymax=0.38,ymin=0,title=1\%
]
\draw node[above,xshift=-2mm,yshift=21mm] {\scriptsize$\uparrow$0.95};
\draw node[above,xshift=16mm,yshift=21mm] {\scriptsize$\uparrow$0.64};
\draw node[above,xshift=60mm,yshift=21mm] {\scriptsize$\uparrow$0.63};
\addplot[orange, fill,draw=black, very thick] coordinates {(A,      0.341   )};
\addplot[blue, fill,draw=black, very thick] coordinates {(B,        0.231   )};
\addplot[teal, fill,draw=black, very thick] coordinates {(C,        0.154   )};
\addplot[lime, fill,draw=black, very thick] coordinates {(D,        0.261   )};
\addplot[brown, fill,draw=black, very thick]  coordinates {(E,      0.341   )};
\addplot[gray1, fill,draw=black, very thick]  coordinates {(F,      0.205   )};
\addplot[gray2, fill,draw=black, very thick] coordinates {(G,       0.237   )};
\addplot[gray3, fill,draw=black, very thick] coordinates {(H,       0.188   )};
\addplot[gray4, fill,draw=black, very thick] coordinates {(I,       0.197   )};
\addplot[gray5, fill,draw=black, very thick] coordinates {(J,       0.161   )};
\addplot[gray6, fill,draw=black, very thick] coordinates {(K,       0.211   )};
\addplot[hotmagenta, fill,draw=black, very thick] coordinates {(L,  0.147   )};
\addplot[yellow, fill,draw=black, very thick] coordinates {(M,      0.175   )};
\addplot[red, fill,draw=black, very thick] coordinates {(N,         0.167   )};
\addplot[cyan, fill,draw=black, very thick] coordinates {(O,        0.341   )};
\addplot[green, fill,draw=black, very thick] coordinates {(P,       0.341   )};
\end{groupplot}
\end{tikzpicture}
\caption{Overall rmsAAD over the Samson dataset reported for different training sizes:
\legendsquare{fill=orange}~LMM,
\legendsquare{fill=blue}~SVR,
\legendsquare{fill=teal}~CB-CNN,
\legendsquare{fill=lime}~WS-AE,
\legendsquare{fill=brown}~UnDIP,
\legendsquare{fill=gray1}~MB(1D),
\legendsquare{fill=gray2}~MB(2D),
\legendsquare{fill=gray3}~MB(3D),
\legendsquare{fill=gray4}~MB(1D+2D),
\legendsquare{fill=gray5}~MB(1D+3D),
\legendsquare{fill=gray6}~MB(2D+3D),
\legendsquare{fill=hotmagenta}~MB,
\legendsquare{fill=yellow}~MB-DR,
\legendsquare{fill=red}~MB-Res,
\legendsquare{fill=cyan}~MB-TL,
\legendsquare{fill=green}~MB-PT. For some methods, we indicate the exact value of rmsAAD above the arrow to maintain readability of the plot (those values are outside the current rmsAAD range on the Y axis).}
\label{fig:rmsAAD_samson}
\end{figure}

\begin{figure}[ht!]
\pgfplotsset{every axis title/.append style={at={(0.5,0.9)}}}
\center
\begin{tikzpicture}[scale=0.8]		
[font=\footnotesize\sffamily]
\begin{groupplot}[
group style={group size=2 by 3, horizontal sep=0.8cm, vertical sep=0.45cm, group name=my plots},
legend style={at={(0.5,-0.15)},
anchor=north,legend columns=-1},
symbolic x coords={A,B,C,D,E,F,G,H,I,J,K,L,M,N,O,P},
x=0.42cm,xticklabels=\empty,
height=4cm,xtick style={draw=none},
ymajorgrids,y tick label style={/pgf/number format/.cd,fixed,fixed zerofill,precision=2,/tikz/.cd}]
\nextgroupplot[
ylabel={rmsAAD},
ybar=-5pt,
bar width=0.2cm,
xtick=data,ymax=0.205,ymin=0,title=100\%
]
\draw node[above,xshift=-2mm,yshift=20.5mm] {\scriptsize$\uparrow$0.62};
\addplot[orange, fill,draw=black, very thick] coordinates {(A,      0.182   )};
\addplot[blue, fill,draw=black, very thick] coordinates {(B,        0.182   )};
\addplot[teal, fill,draw=black, very thick] coordinates {(C,        0.071   )};
\addplot[lime, fill,draw=black, very thick] coordinates {(D,        0.103   )};
\addplot[brown, fill,draw=black, very thick]  coordinates {(E,      0.085   )};
\addplot[gray1, fill,draw=black, very thick]  coordinates {(F,      0.043   )};
\addplot[gray2, fill,draw=black, very thick] coordinates {(G,       0.069   )};
\addplot[gray3, fill,draw=black, very thick] coordinates {(H,       0.043   )};
\addplot[gray4, fill,draw=black, very thick] coordinates {(I,       0.044   )};
\addplot[gray5, fill,draw=black, very thick] coordinates {(J,       0.040   )};
\addplot[gray6, fill,draw=black, very thick] coordinates {(K,       0.041   )};
\addplot[hotmagenta, fill,draw=black, very thick] coordinates {(L,  0.040   )};
\addplot[yellow, fill,draw=black, very thick] coordinates {(M,      0.042   )};
\addplot[red, fill,draw=black, very thick] coordinates {(N,         0.039   )};
\addplot[cyan, fill,draw=black, very thick] coordinates {(O,        0.042   )};
\addplot[green, fill,draw=black, very thick] coordinates {(P,       0.040   )};

\nextgroupplot[
ybar=-5pt,
bar width=0.2cm,
xtick=data,ymax=0.21,ymin=0,title=66\%
]
\draw node[above,xshift=-2mm,yshift=21mm] {\scriptsize$\uparrow$0.62};
\addplot[orange, fill,draw=black, very thick] coordinates {(A,      0.191   )};
\addplot[blue, fill,draw=black, very thick] coordinates {(B,        0.191   )};
\addplot[teal, fill,draw=black, very thick] coordinates {(C,        0.078   )};
\addplot[lime, fill,draw=black, very thick] coordinates {(D,        0.112   )};
\addplot[brown, fill,draw=black, very thick]  coordinates {(E,      0.093   )};
\addplot[gray1, fill,draw=black, very thick]  coordinates {(F,      0.050   )};
\addplot[gray2, fill,draw=black, very thick] coordinates {(G,       0.079   )};
\addplot[gray3, fill,draw=black, very thick] coordinates {(H,       0.049   )};
\addplot[gray4, fill,draw=black, very thick] coordinates {(I,       0.050   )};
\addplot[gray5, fill,draw=black, very thick] coordinates {(J,       0.047   )};
\addplot[gray6, fill,draw=black, very thick] coordinates {(K,       0.047   )};
\addplot[hotmagenta, fill,draw=black, very thick] coordinates {(L,  0.047   )};
\addplot[yellow, fill,draw=black, very thick] coordinates {(M,      0.050   )};
\addplot[red, fill,draw=black, very thick] coordinates {(N,         0.046   )};
\addplot[cyan, fill,draw=black, very thick] coordinates {(O,        0.049   )};
\addplot[green, fill,draw=black, very thick] coordinates {(P,       0.045   )};

\nextgroupplot[
ylabel={rmsAAD},
ybar=-5pt,
bar width=0.2cm,
xtick=data,ymax=0.235,ymin=0,title=33\%
]
\draw node[above,xshift=-2mm,yshift=20.5mm] {\scriptsize$\uparrow$0.62};
\addplot[orange, fill,draw=black, very thick] coordinates {(A,      0.207   )};
\addplot[blue, fill,draw=black, very thick] coordinates {(B,        0.207   )};
\addplot[teal, fill,draw=black, very thick] coordinates {(C,        0.093   )};
\addplot[lime, fill,draw=black, very thick] coordinates {(D,        0.125   )};
\addplot[brown, fill,draw=black, very thick]  coordinates {(E,      0.118   )};
\addplot[gray1, fill,draw=black, very thick]  coordinates {(F,      0.068   )};
\addplot[gray2, fill,draw=black, very thick] coordinates {(G,       0.103   )};
\addplot[gray3, fill,draw=black, very thick] coordinates {(H,       0.068   )};
\addplot[gray4, fill,draw=black, very thick] coordinates {(I,       0.067   )};
\addplot[gray5, fill,draw=black, very thick] coordinates {(J,       0.064   )};
\addplot[gray6, fill,draw=black, very thick] coordinates {(K,       0.067   )};
\addplot[hotmagenta, fill,draw=black, very thick] coordinates {(L,  0.064   )};
\addplot[yellow, fill,draw=black, very thick] coordinates {(M,      0.065   )};
\addplot[red, fill,draw=black, very thick] coordinates {(N,         0.064   )};
\addplot[cyan, fill,draw=black, very thick] coordinates {(O,        0.070   )};
\addplot[green, fill,draw=black, very thick] coordinates {(P,       0.063   )};

\nextgroupplot[
ybar=-5pt,
bar width=0.2cm,
xtick=data,ymax=0.265,ymin=0,title=13\%
]
\draw node[above,xshift=-2mm,yshift=20.5mm] {\scriptsize$\uparrow$0.62};
\addplot[orange, fill,draw=black, very thick] coordinates {(A,      0.234   )};
\addplot[blue, fill,draw=black, very thick] coordinates {(B,        0.234   )};
\addplot[teal, fill,draw=black, very thick] coordinates {(C,        0.127   )};
\addplot[lime, fill,draw=black, very thick] coordinates {(D,        0.145   )};
\addplot[brown, fill,draw=black, very thick]  coordinates {(E,      0.138   )};
\addplot[gray1, fill,draw=black, very thick]  coordinates {(F,      0.102   )};
\addplot[gray2, fill,draw=black, very thick] coordinates {(G,       0.153   )};
\addplot[gray3, fill,draw=black, very thick] coordinates {(H,       0.140   )};
\addplot[gray4, fill,draw=black, very thick] coordinates {(I,       0.102   )};
\addplot[gray5, fill,draw=black, very thick] coordinates {(J,       0.101   )};
\addplot[gray6, fill,draw=black, very thick] coordinates {(K,       0.106   )};
\addplot[hotmagenta, fill,draw=black, very thick] coordinates {(L,  0.102   )};
\addplot[yellow, fill,draw=black, very thick] coordinates {(M,      0.099   )};
\addplot[red, fill,draw=black, very thick] coordinates {(N,         0.103   )};
\addplot[cyan, fill,draw=black, very thick] coordinates {(O,        0.138   )};
\addplot[green, fill,draw=black, very thick] coordinates {(P,       0.096   )};

\nextgroupplot[
ylabel={rmsAAD},
ybar=-5pt,
bar width=0.2cm,
xtick=data,ymax=0.28,title=6\%
]
\draw node[above,xshift=-2mm,yshift=18.5mm] {\scriptsize$\uparrow$0.63};
\draw node[above,xshift=60mm,yshift=18.5mm] {\scriptsize$\uparrow$0.30};
\addplot[orange, fill,draw=black, very thick] coordinates {(A,      0.261   )};
\addplot[blue, fill,draw=black, very thick] coordinates {(B,        0.261   )};
\addplot[teal, fill,draw=black, very thick] coordinates {(C,        0.160   )};
\addplot[lime, fill,draw=black, very thick] coordinates {(D,        0.176   )};
\addplot[brown, fill,draw=black, very thick]  coordinates {(E,      0.170   )};
\addplot[gray1, fill,draw=black, very thick]  coordinates {(F,      0.134   )};
\addplot[gray2, fill,draw=black, very thick] coordinates {(G,       0.196   )};
\addplot[gray3, fill,draw=black, very thick] coordinates {(H,       0.102   )};
\addplot[gray4, fill,draw=black, very thick] coordinates {(I,       0.137   )};
\addplot[gray5, fill,draw=black, very thick] coordinates {(J,       0.138   )};
\addplot[gray6, fill,draw=black, very thick] coordinates {(K,       0.142   )};
\addplot[hotmagenta, fill,draw=black, very thick] coordinates {(L,  0.138   )};
\addplot[yellow, fill,draw=black, very thick] coordinates {(M,      0.133   )};
\addplot[red, fill,draw=black, very thick] coordinates {(N,         0.140   )};
\addplot[cyan, fill,draw=black, very thick] coordinates {(O,        0.261   )};
\addplot[green, fill,draw=black, very thick] coordinates {(P,       0.133   )};

\nextgroupplot[
ybar=-5pt,
bar width=0.2cm,
xtick=data,ymax=0.51,ymin=0,title=1\%
]
\draw node[above,xshift=-2mm,yshift=17mm] {\scriptsize$\uparrow$0.69};
\draw node[above,xshift=60mm,yshift=17mm] {\scriptsize$\uparrow$1.09};
\addplot[orange, fill,draw=black, very thick] coordinates {(A,      0.360   )};
\addplot[blue, fill,draw=black, very thick] coordinates {(B,        0.360   )};
\addplot[teal, fill,draw=black, very thick] coordinates {(C,        0.312   )};
\addplot[lime, fill,draw=black, very thick] coordinates {(D,        0.321   )};
\addplot[brown, fill,draw=black, very thick]  coordinates {(E,      0.493   )};
\addplot[gray1, fill,draw=black, very thick]  coordinates {(F,      0.282   )};
\addplot[gray2, fill,draw=black, very thick] coordinates {(G,       0.328   )};
\addplot[gray3, fill,draw=black, very thick] coordinates {(H,       0.298   )};
\addplot[gray4, fill,draw=black, very thick] coordinates {(I,       0.279   )};
\addplot[gray5, fill,draw=black, very thick] coordinates {(J,       0.281   )};
\addplot[gray6, fill,draw=black, very thick] coordinates {(K,       0.298   )};
\addplot[hotmagenta, fill,draw=black, very thick] coordinates {(L,  0.279   )};
\addplot[yellow, fill,draw=black, very thick] coordinates {(M,      0.284   )};
\addplot[red, fill,draw=black, very thick] coordinates {(N,         0.287   )};
\addplot[cyan, fill,draw=black, very thick] coordinates {(O,        0.360   )};
\addplot[green, fill,draw=black, very thick] coordinates {(P,       0.458   )};
\end{groupplot}
\end{tikzpicture}
\caption{Overall rmsAAD over the Urban dataset reported for different training sizes:
\legendsquare{fill=orange}~LMM,
\legendsquare{fill=blue}~SVR,
\legendsquare{fill=teal}~CB-CNN,
\legendsquare{fill=lime}~WS-AE,
\legendsquare{fill=brown}~UnDIP,
\legendsquare{fill=gray1}~MB(1D),
\legendsquare{fill=gray2}~MB(2D),
\legendsquare{fill=gray3}~MB(3D),
\legendsquare{fill=gray4}~MB(1D+2D),
\legendsquare{fill=gray5}~MB(1D+3D),
\legendsquare{fill=gray6}~MB(2D+3D),
\legendsquare{fill=hotmagenta}~MB,
\legendsquare{fill=yellow}~MB-DR,
\legendsquare{fill=red}~MB-Res,
\legendsquare{fill=cyan}~MB-TL,
\legendsquare{fill=green}~MB-PT. For some methods, we indicate the exact value of rmsAAD above the arrow to maintain readability of the plot (those values are outside the current rmsAAD range on the Y axis).}
\label{fig:rmsaad_urban}
\end{figure}

\begin{figure}[ht!]
\pgfplotsset{every axis title/.append style={at={(0.5,0.9)}}}
\center
\begin{tikzpicture}[scale=0.8]		
[font=\footnotesize\sffamily]
\begin{groupplot}[
group style={group size=2 by 3, horizontal sep=0.8cm, vertical sep=0.45cm, group name=my plots},
legend style={at={(0.5,-0.15)},
anchor=north,legend columns=-1},
symbolic x coords={A,B,C,D,E,F,G,H,I,J,K,L,M,N,O,P},
x=0.42cm,xticklabels=\empty,
height=4cm,xtick style={draw=none},
ymajorgrids,y tick label style={/pgf/number format/.cd,fixed,fixed zerofill,precision=2,/tikz/.cd}]
\nextgroupplot[
ylabel={rmsAAD},
ybar=-5pt,
bar width=0.2cm,
xtick=data,ymax=0.13,ymin=0,title=100\%
]
\draw node[above,xshift=-2mm,yshift=20mm] {\scriptsize$\uparrow$0.19};
\draw node[above,xshift=60mm,yshift=20mm] {\scriptsize$\uparrow$0.14};
\addplot[orange, fill,draw=black, very thick] coordinates {(A,      0.112   )};
\addplot[blue, fill,draw=black, very thick] coordinates {(B,        0.112   )};
\addplot[teal, fill,draw=black, very thick] coordinates {(C,        0.052   )};
\addplot[lime, fill,draw=black, very thick] coordinates {(D,        0.066   )};
\addplot[brown, fill,draw=black, very thick]  coordinates {(E,      0.068   )};
\addplot[gray1, fill,draw=black, very thick]  coordinates {(F,      0.037   )};
\addplot[gray2, fill,draw=black, very thick] coordinates {(G,       0.046   )};
\addplot[gray3, fill,draw=black, very thick] coordinates {(H,       0.040   )};
\addplot[gray4, fill,draw=black, very thick] coordinates {(I,       0.036   )};
\addplot[gray5, fill,draw=black, very thick] coordinates {(J,       0.034   )};
\addplot[gray6, fill,draw=black, very thick] coordinates {(K,       0.038   )};
\addplot[hotmagenta, fill,draw=black, very thick] coordinates {(L,  0.035   )};
\addplot[yellow, fill,draw=black, very thick] coordinates {(M,      0.035   )};
\addplot[red, fill,draw=black, very thick] coordinates {(N,         0.036   )};
\addplot[cyan, fill,draw=black, very thick] coordinates {(O,        0.112   )};
\addplot[green, fill,draw=black, very thick] coordinates {(P,       0.059   )};

\nextgroupplot[
ybar=-5pt,
bar width=0.2cm,
xtick=data,ymax=0.135,ymin=0,title=66\%
]
\draw node[above,xshift=-2mm,yshift=20mm] {\scriptsize$\uparrow$0.19};
\draw node[above,xshift=12mm,yshift=20mm] {\scriptsize$\uparrow$0.14};
\draw node[above,xshift=60mm,yshift=20mm] {\scriptsize$\uparrow$0.27};
\addplot[orange, fill,draw=black, very thick] coordinates {(A,      0.117   )};
\addplot[blue, fill,draw=black, very thick] coordinates {(B,        0.117   )};
\addplot[teal, fill,draw=black, very thick] coordinates {(C,        0.097   )};
\addplot[lime, fill,draw=black, very thick] coordinates {(D,        0.117   )};
\addplot[brown, fill,draw=black, very thick]  coordinates {(E,      0.073   )};
\addplot[gray1, fill,draw=black, very thick]  coordinates {(F,      0.046   )};
\addplot[gray2, fill,draw=black, very thick] coordinates {(G,       0.054   )};
\addplot[gray3, fill,draw=black, very thick] coordinates {(H,       0.047   )};
\addplot[gray4, fill,draw=black, very thick] coordinates {(I,       0.043   )};
\addplot[gray5, fill,draw=black, very thick] coordinates {(J,       0.042   )};
\addplot[gray6, fill,draw=black, very thick] coordinates {(K,       0.044   )};
\addplot[hotmagenta, fill,draw=black, very thick] coordinates {(L,  0.041   )};
\addplot[yellow, fill,draw=black, very thick] coordinates {(M,      0.042   )};
\addplot[red, fill,draw=black, very thick] coordinates {(N,         0.042   )};
\addplot[cyan, fill,draw=black, very thick] coordinates {(O,        0.117   )};
\addplot[green, fill,draw=black, very thick] coordinates {(P,       0.040   )};
\nextgroupplot[
ylabel={rmsAAD},
ybar=-5pt,
bar width=0.2cm,
xtick=data,ymax=0.145,ymin=0,title=33\%
]
\draw node[above,xshift=-2mm,yshift=20.5mm] {\scriptsize$\uparrow$0.19};
\draw node[above,xshift=12mm,yshift=20.5mm] {\scriptsize$\uparrow$0.24};
\draw node[above,xshift=60mm,yshift=20.5mm] {\scriptsize$\uparrow$0.49};
\addplot[orange, fill,draw=black, very thick] coordinates {(A,      0.128   )};
\addplot[blue, fill,draw=black, very thick] coordinates {(B,        0.128   )};
\addplot[teal, fill,draw=black, very thick] coordinates {(C,        0.113   )};
\addplot[lime, fill,draw=black, very thick] coordinates {(D,        0.128   )};
\addplot[brown, fill,draw=black, very thick]  coordinates {(E,      0.088   )};
\addplot[gray1, fill,draw=black, very thick]  coordinates {(F,      0.064   )};
\addplot[gray2, fill,draw=black, very thick] coordinates {(G,       0.072   )};
\addplot[gray3, fill,draw=black, very thick] coordinates {(H,       0.064   )};
\addplot[gray4, fill,draw=black, very thick] coordinates {(I,       0.057   )};
\addplot[gray5, fill,draw=black, very thick] coordinates {(J,       0.060   )};
\addplot[gray6, fill,draw=black, very thick] coordinates {(K,       0.061   )};
\addplot[hotmagenta, fill,draw=black, very thick] coordinates {(L,  0.055   )};
\addplot[yellow, fill,draw=black, very thick] coordinates {(M,      0.055   )};
\addplot[red, fill,draw=black, very thick] coordinates {(N,         0.060   )};
\addplot[cyan, fill,draw=black, very thick] coordinates {(O,        0.128   )};
\addplot[green, fill,draw=black, very thick] coordinates {(P,       0.081   )};
\nextgroupplot[
ybar=-5pt,
bar width=0.2cm,
xtick=data,ymax=0.355,ymin=0,title=13\%
]
\draw node[above,xshift=60mm,yshift=20.5mm] {\scriptsize$\uparrow$0.74};
\addplot[orange, fill,draw=black, very thick] coordinates {(A,      0.195   )};
\addplot[blue, fill,draw=black, very thick] coordinates {(B,        0.153   )};
\addplot[teal, fill,draw=black, very thick] coordinates {(C,        0.148   )};
\addplot[lime, fill,draw=black, very thick] coordinates {(D,        0.250   )};
\addplot[brown, fill,draw=black, very thick]  coordinates {(E,      0.125   )};
\addplot[gray1, fill,draw=black, very thick]  coordinates {(F,      0.092   )};
\addplot[gray2, fill,draw=black, very thick] coordinates {(G,       0.112   )};
\addplot[gray3, fill,draw=black, very thick] coordinates {(H,       0.137   )};
\addplot[gray4, fill,draw=black, very thick] coordinates {(I,       0.086   )};
\addplot[gray5, fill,draw=black, very thick] coordinates {(J,       0.110   )};
\addplot[gray6, fill,draw=black, very thick] coordinates {(K,       0.109   )};
\addplot[hotmagenta, fill,draw=black, very thick] coordinates {(L,  0.102   )};
\addplot[yellow, fill,draw=black, very thick] coordinates {(M,      0.082   )};
\addplot[red, fill,draw=black, very thick] coordinates {(N,         0.116   )};
\addplot[cyan, fill,draw=black, very thick] coordinates {(O,        0.317   )};
\addplot[green, fill,draw=black, very thick] coordinates {(P,       0.317   )};
\nextgroupplot[
ylabel={rmsAAD},
ybar=-5pt,
bar width=0.2cm,
xtick=data,ymax=0.23,title=6\%
]
\draw node[above,xshift=59mm,yshift=18.5mm] {\scriptsize$\uparrow$0.79};
\draw node[above,xshift=65mm,yshift=18.5mm] {\scriptsize$\uparrow$0.58};
\draw node[above,xshift=42,yshift=18.5mm] {\scriptsize$\uparrow$0.37};
\addplot[orange, fill,draw=black, very thick] coordinates {(A,      0.215   )};
\addplot[blue, fill,draw=black, very thick] coordinates {(B,        0.177   )};
\addplot[teal, fill,draw=black, very thick] coordinates {(C,        0.209   )};
\addplot[lime, fill,draw=black, very thick] coordinates {(D,        0.172   )};
\addplot[brown, fill,draw=black, very thick]  coordinates {(E,      0.215   )};
\addplot[gray1, fill,draw=black, very thick]  coordinates {(F,      0.119   )};
\addplot[gray2, fill,draw=black, very thick] coordinates {(G,       0.162   )};
\addplot[gray3, fill,draw=black, very thick] coordinates {(H,       0.125   )};
\addplot[gray4, fill,draw=black, very thick] coordinates {(I,       0.109   )};
\addplot[gray5, fill,draw=black, very thick] coordinates {(J,       0.128   )};
\addplot[gray6, fill,draw=black, very thick] coordinates {(K,       0.131   )};
\addplot[hotmagenta, fill,draw=black, very thick] coordinates {(L,  0.109   )};
\addplot[yellow, fill,draw=black, very thick] coordinates {(M,      0.108   )};
\addplot[red, fill,draw=black, very thick] coordinates {(N,         0.119   )};
\addplot[cyan, fill,draw=black, very thick] coordinates {(O,        0.215   )};
\addplot[green, fill,draw=black, very thick] coordinates {(P,       0.215   )};
\nextgroupplot[
ybar=-5pt,
bar width=0.2cm,
xtick=data,ymax=0.475,ymin=0,title=1\%
]
\draw node[above,xshift=59mm,yshift=20.5mm] {\scriptsize$\uparrow$0.89};
\draw node[above,xshift=65mm,yshift=20.5mm] {\scriptsize$\uparrow$0.84};
\draw node[above,xshift=42,yshift=20.5mm] {\scriptsize$\uparrow$0.71};
\addplot[orange, fill,draw=black, very thick] coordinates {(A,      0.289   )};
\addplot[blue, fill,draw=black, very thick] coordinates {(B,        0.285   )};
\addplot[teal, fill,draw=black, very thick] coordinates {(C,        0.361   )};
\addplot[lime, fill,draw=black, very thick] coordinates {(D,        0.422   )};
\addplot[brown, fill,draw=black, very thick]  coordinates {(E,      0.422   )};
\addplot[gray1, fill,draw=black, very thick]  coordinates {(F,      0.360   )};
\addplot[gray2, fill,draw=black, very thick] coordinates {(G,       0.352   )};
\addplot[gray3, fill,draw=black, very thick] coordinates {(H,       0.355   )};
\addplot[gray4, fill,draw=black, very thick] coordinates {(I,       0.340   )};
\addplot[gray5, fill,draw=black, very thick] coordinates {(J,       0.318   )};
\addplot[gray6, fill,draw=black, very thick] coordinates {(K,       0.370   )};
\addplot[hotmagenta, fill,draw=black, very thick] coordinates {(L,  0.339   )};
\addplot[yellow, fill,draw=black, very thick] coordinates {(M,      0.331   )};
\addplot[red, fill,draw=black, very thick] coordinates {(N,         0.334   )};
\addplot[cyan, fill,draw=black, very thick] coordinates {(O,        0.422   )};
\addplot[green, fill,draw=black, very thick] coordinates {(P,       0.422   )};

\end{groupplot}
\end{tikzpicture}
\caption{Overall rmsAAD over the Jasper Ridge dataset reported for different training sizes:
\legendsquare{fill=orange}~LMM,
\legendsquare{fill=blue}~SVR,
\legendsquare{fill=teal}~CB-CNN,
\legendsquare{fill=lime}~WS-AE,
\legendsquare{fill=brown}~UnDIP,
\legendsquare{fill=gray1}~MB(1D),
\legendsquare{fill=gray2}~MB(2D),
\legendsquare{fill=gray3}~MB(3D),
\legendsquare{fill=gray4}~MB(1D+2D),
\legendsquare{fill=gray5}~MB(1D+3D),
\legendsquare{fill=gray6}~MB(2D+3D),
\legendsquare{fill=hotmagenta}~MB,
\legendsquare{fill=yellow}~MB-DR,
\legendsquare{fill=red}~MB-Res,
\legendsquare{fill=cyan}~MB-TL,
\legendsquare{fill=green}~MB-PT. For some methods, we indicate the exact value of rmsAAD above the arrow to maintain readability of the plot (those values are outside the current rmsAAD range on the Y axis).}
\label{fig:rmsaad_jasper}
\end{figure}

\begin{table*}[ht!]
\caption{The ranking (RMSE and rmsAAD) obtained for the Samson dataset by all HU algorithms. The best ranking for each training set size is boldfaced, whereas the second best is underlined.}\label{tab:ranking_samson}
\centering
\setlength{\tabcolsep}{1mm}
\scalebox{0.7}{
\begin{tabular}{r@{\hskip 2mm}|@{\hskip 2mm}ccccccc@{\hskip 3mm}|@{\hskip 3mm}ccccccc}
\Xhline{2\arrayrulewidth}
\multicolumn{1}{c}{}              & \multicolumn{7}{c}{\textbf{RMSE}}                                                                                                                                 & \multicolumn{7}{c}{\textbf{rmsAAD}}                                                                                                                   \\
\Xhline{1.5\arrayrulewidth}
\textbf{Train. size $\rightarrow$}                         & \textbf{100\%} & \textbf{66\%} & \textbf{33\%} & \textbf{13\%} & \textbf{6\%} & \textbf{1\%} & \textbf{Mean}              & \textbf{100\%} & \textbf{66\%} & \textbf{33\%} & \textbf{13\%} & \textbf{6\%} & \textbf{1\%} & \textbf{Mean}  \\\Xhline{1.5\arrayrulewidth}
LMM               &16.000&	16.000      &16.000     &16.000     &16.000&	16.000	&16.000	&16.000&	16.000&	16.000&	16.000&	16.000&	16.000  &	16.000\\
SVR                      & 15.000	&15.000	    &13.000     &3.000   &13.000&	10.000&	13.167&	15.000&	15.000	&13.000&	13.000&	13.000&	10.000  &	13.167\\
CB-CNN              & 12.000	&10.000&	10.000	    &9.000      &9.000&	{\ul2.000}&	8.667&	12.000&	10.000&	10.000&	9.000&	9.000	&{\ul2.000	  }      &8.667 \\
WS-AE                   & 13.000	&11.000&	12.000      &10.000     &10.000&	12.000&	11.333&	13.000&	11.000&	11.000&	10.000	&10.000&	12.000  &	11.167\\
UnDIP             &14.000	    &14.000&	14.000      &15.000     &14.000	&14.000&	14.167	&14.000	&14.000	&14.000&	14.000	&14.000&	15.000&	14.167\\ \Xhline{0.5\arrayrulewidth}
MB(1D)                                  & 8.000	    &7.500      &7.500      &6.500      &5.500&	9.000&	7.333&	7.500&	7.000&	7.500&	6.000&	5.000&	8.000           &	6.833\\
MB(2D)                                  &11.000	    &9.000	    &11.000	    &11.000	    &11.000	&11.000&	10.667&	11.000&	9.000&	12.000&	12.000&	11.000&	11.000      &	11.000\\
MB(3D)                                  &9.000	    &7.500	    &9.000	    &8.000	    &8.000	&6.000	&7.917	&9.000	&8.000&	9.000&	8.000&	8.000&	6.000	        &8.000\\ \Xhline{0.5\arrayrulewidth}
MB(1D+2D)                               &{\ul4.000}	    &{\ul2.500}	    &4.500	    &5.000      &5.500&	7.000&	4.750	&5.000	&4.500&	5.500&	5.000&	6.000&	7.000   &	5.500 \\
MB(1D+3D)                               &{\ul4.000}	    &6.000	    &4.500	    &{\ul2.000}	    &{\ul2.000}	&3.000&	{\ul3.583}&	5.000&	4.500&	5.500	&\textbf{1.000}	&{\ul2.000}	&3.000	&3.500 \\
MB(2D+3D)                               &7.000	    &{\ul2.500}	    &7.500	    &6.500	    &7.000&	8.000	&6.417	&7.500	&6.000	&7.500&	7.000&	7.000&	9.000&	7.333 \\\Xhline{0.5\arrayrulewidth}
\textbf{MB}                             & {\ul4.000}	    &\textbf{1.000}	    &\textbf{1.000}	    &\textbf{1.000}	    &\textbf{1.000}	&\textbf{1.000}	&\textbf{1.500}	&{\ul2.500}	&\textbf{1.000}	&\textbf{1.000}	&{\ul2.000}	&\textbf{1.000}	&\textbf{1.000}	&\textbf{1.417}\\
\textbf{MB-DR}                          &{\ul4.000}	    &4.000	    &6.000	    &3.500	    &4.000	&5.000	&4.417	&5.000	&{\ul2.000}	&3.000	&3.000	&4.000	&5.000	&3.667\\
\textbf{MB-Res}                         & {\ul4.000}	    &5.000	    &{\ul2.000}	    &3.500	    &3.000	&4.000	&{\ul3.583}	&{\ul2.500}	&3.000	&4.000	&4.000	&3.000	&4.000	&{\ul3.417}\\
\textbf{MB-TL}                          &10.000	    &13.000	    &15.000	    &14.000	    &15.000	&15.000	&13.667	&10.000	&13.000	&15.000	&15.000	&15.000	&14.000	&13.667 \\
\textbf{MB-PT}                          &\textbf{1.000}	    &12.000	    &3.000	    &12.000	    &12.000	&13.000	&8.833	&\textbf{1.000}	&12.000	&{\ul2.000}	&11.000	&12.000	&13.000	&8.500\\\Xhline{2\arrayrulewidth}
\end{tabular}}
\end{table*}

\begin{table*}[ht!]
\caption{The ranking (RMSE and rmsAAD) obtained for the Urban dataset by all HU algorithms. The best ranking for each training set size is boldfaced, whereas the second best is underlined.}\label{tab:ranking_urban}
\centering
\setlength{\tabcolsep}{1mm}
\scalebox{0.7}{
\begin{tabular}{r@{\hskip 2mm}|@{\hskip 2mm}ccccccc@{\hskip 3mm}|@{\hskip 3mm}ccccccc}
\Xhline{2\arrayrulewidth}
\multicolumn{1}{c}{}              & \multicolumn{7}{c}{\textbf{RMSE}}                                                                                                                                 & \multicolumn{7}{c}{\textbf{rmsAAD}}                                                                                                                   \\
\Xhline{1.5\arrayrulewidth}
\textbf{Train. size $\rightarrow$}                         & \textbf{100\%} & \textbf{66\%} & \textbf{33\%} & \textbf{13\%} & \textbf{6\%} & \textbf{1\%} & \textbf{Mean}              & \textbf{100\%} & \textbf{66\%} & \textbf{33\%} & \textbf{13\%} & \textbf{6\%} & \textbf{1\%} & \textbf{Mean}  \\\Xhline{1.5\arrayrulewidth}
LMM       &16.000	&16.000	&16.000	&16.000	&16.000	&15.000	&15.833 &16.000	&16.000	&16.000	&16.000	&16.000	&15.000	&15.833  \\
SVR              &15.000	&15.000	&15.000	&15.000	&14.000	&12.000	&14.333 &15.000	&15.000	&15.000	&15.000	&14.000	&12.000	&14.333  \\
CB-CNN      &12.000	&12.000	&11.000	&9.000	&10.000	&9.000	&10.500 &12.000	&11.000	&11.000	&9.000	&10.000	&9.000	&10.333  \\
WS-AE           &14.000	&14.000	&14.000	&14.000	&12.000	&10.000	&13.000 &14.000	&14.000	&14.000	&13.000	&12.000	&10.000	&12.833  \\
UnDIP     &13.000	&13.000	&13.000	&11.000	&11.000	&14.000	&12.500 &13.000	&13.000	&13.000	&11.000	&11.000	&14.000	&12.500  \\\Xhline{0.5\arrayrulewidth}
MB(1D)                          &6.000	&7.000	&6.500	&4.000	&6.000	&5.000	&5.750  &8.500	&8.500	&8.500	&4.500	&4.000	&4.000	&6.333  \\
MB(2D)                          &11.000	&11.000	&12.000	&13.000	&13.000	&11.000	&11.833 &11.000	&12.000	&12.000	&14.000	&13.000	&11.000	&12.167  \\
MB(3D)                          &10.000	&7.000	&9.500	&12.000	&\textbf{1.000}	&6.500	&7.667  &8.500	&7.000	&8.500	&12.000	&\textbf{1.000}	&7.500	&7.417  \\\Xhline{0.5\arrayrulewidth}
MB(1D+2D)                       &6.000	&7.000	&6.500	&4.000	&6.000	&{\ul2.500}	&5.333  &10.000	&8.500	&6.500	&4.500	&5.000	&\textbf{1.500}	&6.000  \\
MB(1D+3D)                       &6.000	&7.000	&\textbf{1.500}	&4.000	&6.000	&{\ul2.500}	&4.500  &3.500	&4.500	&{\ul2.500}	&3.000	&6.000	&{\ul3.000}	&{\ul3.750}  \\
MB(2D+3D)                       &6.000	&7.000	&9.500	&8.000	&8.000	&6.500	&7.500  &5.000	&4.500	&6.500	&8.000	&9.000	&7.500	&6.750  \\\Xhline{0.5\arrayrulewidth}
\textbf{MB}                     &6.000	&3.000	&{\ul3.000}	&6.000	&4.000	&\textbf{1.000}	&{\ul3.833}  &3.500	&3.000	&{\ul2.500}	&6.000	&7.000	&\textbf{1.500}	&3.917  \\
\textbf{MB-DR}                  &6.000	&10.000	&5.000	&{\ul2.000}	&3.000	&4.000	&5.000  &6.000	&10.000	&5.000	&{\ul2.000}	&{\ul2.000}	&5.000	&5.000  \\
\textbf{MB-Res}                 &\textbf{1.000}	&{\ul2.000}	&4.000	&7.000	&9.000	&8.000	&5.167  &\textbf{1.000}	&{\ul2.000}	&4.000	&7.000	&8.000	&6.000	&4.667  \\
\textbf{MB-TL}                  &6.000	&4.000	&8.000	&10.000	&15.000	&16.000	&9.833  &7.000	&6.000	&10.000	&10.000	&15.000	&16.000	&10.667  \\
\textbf{MB-PT}                  &{\ul2.000}	&\textbf{1.000}	&\textbf{1.500}	&\textbf{1.000}	&{\ul2.000}	&13.000	&\textbf{3.417}  &{\ul2.000}	&\textbf{1.000}	&\textbf{1.000}	&\textbf{1.000}	&3.000	&13.000	&\textbf{3.500}  \\\Xhline{2\arrayrulewidth}
\end{tabular}}
\end{table*}

\begin{table*}[ht!]
\caption{The ranking (RMSE and rmsAAD) obtained for Jasper Ridge by all HU algorithms. The best ranking for each training set size is boldfaced, whereas the second best is underlined.}\label{tab:ranking_jasper}
\centering
\setlength{\tabcolsep}{1mm}
\scalebox{0.7}{
\begin{tabular}{r@{\hskip 2mm}|@{\hskip 2mm}ccccccc@{\hskip 3mm}|@{\hskip 3mm}ccccccc}
\Xhline{2\arrayrulewidth}
\multicolumn{1}{c}{}              & \multicolumn{7}{c}{\textbf{RMSE}}                                                                                                                                 & \multicolumn{7}{c}{\textbf{rmsAAD}}                                                                                                                   \\
\Xhline{1.5\arrayrulewidth}
\textbf{Train. size $\rightarrow$}                         & \textbf{100\%} & \textbf{66\%} & \textbf{33\%} & \textbf{13\%} & \textbf{6\%} & \textbf{1\%} & \textbf{Mean}              & \textbf{100\%} & \textbf{66\%} & \textbf{33\%} & \textbf{13\%} & \textbf{6\%} & \textbf{1\%} & \textbf{Mean}  \\\Xhline{1.5\arrayrulewidth}
LMM       &16.000	        &15.000	&14.000	&13.000	&13.000	&2.000	&12.167 &16.000	&15.000	&14.000	&13.000	&13.000	&{\ul2.000}	&12.167     \\
SVR              &14.000	        &13.000	&13.000	&12.000	&11.000	&1.000	&10.667 &14.000	&13.000	&13.000	&12.000	&11.000	&\textbf{1.000}	&10.667     \\
CB-CNN      &10.000	        &12.000	&12.000	&11.000	&12.000	&11.000	&11.333 &10.000	&12.000	&12.000	&11.000	&12.000	&11.000	&11.333     \\
WS-AE           &12.000	        &14.000	&15.000	&14.000	&10.000	&13.000	&13.000 &12.000	&14.000	&15.000	&14.000	&10.000	&13.000	&13.000     \\
UnDIP     &13.000	        &11.000	&11.000	&9.000	&14.000	&14.000	&12.000 &13.000	&11.000	&11.000	&9.000	&14.000	&14.000	&12.000     \\\Xhline{0.5\arrayrulewidth}
MB(1D)                          & {\ul 6.500}	&9.000	&8.000	&3.000	&5.000	&9.000	&6.750  &6.000	&8.000	&7.500	&3.000	&5.000	&10.000	&6.583   \\
MB(2D)                          &9.000	        &10.000	&9.000	&6.500	&9.000	&8.000	&8.583  &9.000	&10.000	&9.000	&7.000	&9.000	&8.000	&8.667    \\
MB(3D)                          &8.000	        &8.000	&7.000	&10.000	&6.000	&10.000	&8.167  &8.000	&9.000	&7.500	&10.000	&6.000	&9.000	&8.250   \\\Xhline{0.5\arrayrulewidth}
MB(1D+2D)                       &\textbf{3.000}	&6.000	&3.000	&{\ul2.000}	&3.000	&6.000	&3.833  &4.500	&6.000	&3.000	&{\ul2.000}	&3.000	&7.000	&4.250   \\
MB(1D+3D)                       &\textbf{3.000}	&6.000	&5.500	&6.500	&7.000	&3.000	&5.167  &\textbf{1.000}	&5.000	&5.000	&6.000	&7.000	&3.000	&4.500   \\
MB(2D+3D)                       &{\ul6.500}	    &6.000	&5.500	&5.000	&8.000	&12.000	&7.167  &7.000	&7.000	&6.000	&5.000	&8.000	&12.000	&7.500   \\\Xhline{0.5\arrayrulewidth}
\textbf{MB}                     &\textbf{3.000}	&{\ul2.000}	&\textbf{1.000}	&4.000	&{\ul2.000}	&7.000	&{\ul3.167}  &{\ul2.500}	&{\ul2.000}	&\textbf{1.000}	&4.000	&{\ul2.000}	&6.000	&{\ul2.917}   \\
\textbf{MB-DR}                  &\textbf{3.000}	&3.500	&{\ul2.000}	&\textbf{1.000}	&\textbf{1.000}	&5.000	&\textbf{2.583}  &{\ul2.500}	&3.000	&{\ul2.000}	&\textbf{1.000}	&\textbf{1.000}	&4.000	&\textbf{2.250}   \\
\textbf{MB-Res}                 &\textbf{3.000}	&3.500	&4.000	&8.000	&4.000	&4.000	&4.417  &4.500	&4.000	&4.000	&8.000	&4.000	&5.000	&4.917   \\
\textbf{MB-TL}                  &15.000	        &16.000	&16.000	&16.000	&16.000	&16.000	&15.833 &15.000	&16.000	&16.000	&16.000	&16.000	&16.000	&15.833     \\
\textbf{MB-PT}                  &11.000	        &\textbf{1.000}	&10.000	&15.000	&15.000	&15.000	&11.167 &11.000	&\textbf{1.000}	&10.000	&15.000	&15.000	&15.000	&11.167    \\\Xhline{2\arrayrulewidth}
\end{tabular}}
\end{table*}

\begin{table}[ht!]
\scriptsize
\centering
\resizebox{1\textwidth}{!}{
\renewcommand{\tabcolsep}{1mm}
\caption{The impact of the white zero-mean Gaussian noise added to the original data (with the signal-to-noise ratio of 20, 30, 40, and 50 dB) on the performance (quantified as mean, median, and standard deviation of RMSE) of all investigated unmixing algorithms over Jasper Ridge.}\label{tab:noise_impact}
\begin{tabular}{rrrrrrrrrrrrrrrrrrrr}
\Xhline{2\arrayrulewidth}
          &                \multicolumn{3}{c}{\textbf{Original}}                                     &                 &                      & \textbf{50 dB}                        &                      &                 &                      & \textbf{40 dB}                        &                &           &                      & \textbf{30 dB}                        &                &           &                & \textbf{20 dB}         &                \\
          \cline{2-4} \cline{6-8} \cline{10-12} \cline{14-16} \cline{18-20}
        \textbf{Models}  & \textbf{Mean}           & \multicolumn{1}{c}{$\sigma$}                    & \textbf{Med.}         &                 & \textbf{Mean}                 & \multicolumn{1}{c}{$\sigma$}                    & \textbf{Med.}               &                 & \textbf{Mean}                 & \multicolumn{1}{c}{$\sigma$}                    & \textbf{Med.}         &           & \textbf{Mean}                 & \multicolumn{1}{c}{$\sigma$}                    & \textbf{Med.}         &           & \textbf{Mean}           & \multicolumn{1}{c}{$\sigma$}      & \textbf{Med.}         \\
          \hline
LMM       & 0.078          & {0.001}                  & 0.078          &                 & {0.078}          & {0.001}                  & 0.078                & {}          & 0.078                & {0.001}                  & {0.078}    &           & 0.078                & {{0.001}}         & 0.078          &           & 0.078          & {0.001} & 0.078          \\
SVR      & 0.051          & {0.001}                  & 0.051          &                 & {0.051}          & {0.001}                  & 0.051                & {}          & 0.051                & {0.001}                  & {0.051}    &           & 0.051                & {0.001}                  & 0.051          &           & 0.054          & {0.002}    & 0.054          \\
CB-CNN    & 0.021          & 0.002                        & 0.020          &                 & 0.021                & 0.002                        & 0.020                &                 & 0.021                & 0.002                        & 0.020          &           & 0.021                & 0.002                        & 0.021          &           & 0.026          & 0.004          & 0.025          \\
WS-AE    & 0.027          & {0.000} & 0.026          &                 & {0.027}       & {0.000} & 0.026                & {}       & 0.027                & {0.000} & {0.026} &           & 0.027                & {0.000} & 0.027          &           & 0.031          & {0.001} & 0.030          \\
UnDIP     & 0.028          & 0.004                        & 0.027          &                 & 0.028                & 0.004                        & 0.027                &                 & 0.028                & 0.004                        & 0.027          &           & 0.028                & 0.004                        & 0.027          &           & 0.030          & 0.004          & 0.029          \\
MB(1D)    & {0.015}    & {0.001}                  & 0.015          & {}          & {0.015}          & {0.001}                  & {0.015}          & {}          & {0.015}          & {0.001}                  & {0.015}    &           & 0.017                & {0.001}                  & 0.016          &           & 0.027          & 0.004          & 0.026          \\
MB(2D)    & 0.018          & {0.001}                  & 0.018          &                 & {0.018}          & {0.001}                  & 0.018                & {}          & 0.018                & {0.001}                  & {0.018}    &           & 0.019                & {{0.001}}         & 0.019          &           & 0.027          & {0.001} & 0.027          \\
MB(3D)    & 0.016          & {0.001}                  & 0.016          &                 & {0.016}          & {0.001}                  & 0.016                & {}          & 0.016                & {0.001}                  & {0.016}    &           & {0.017}       & {{0.001}}         & {0.016} &           & {0.021} & {0.001} & {0.021} \\
MB(1D+2D) & {0.014} & {0.001}                  & {0.014}    & {{}} & {{0.014}} & {0.001}                  & {{0.014}} & {}          & {{0.014}} & {0.001}                  & {0.014}    & {}    & {0.016}          & {0.001}                  & 0.016          &           & 0.028          & 0.005          & 0.026          \\
MB(1D+3D) & {0.014} & {0.001}                  & {0.014}    & {{}} & {{0.014}} & {0.001}                  & {{0.014}} & {}          & {{0.014}} & {{0.001}}         & {0.014}    & {}    & {0.015}       & {0.001}                  & {0.015}    & {}    & 0.024          & 0.004          & 0.023          \\
MB(2D+3D) & {0.015}    & {0.001}                  & 0.015          & {}          & {0.015}          & {0.001}                  & {0.015}          & {}          & {0.015}          & {0.001}                  & {0.015}    &           & {{0.016}} & {{0.001}}         & {0.016} &           & {0.021} & {0.001} & {0.021} \\
MB        & {0.014} & {0.001}                  & {0.013} & {}       & {{0.014}} & {{0.001}}         & {0.013}       & {{}} & {{0.014}} & {{0.001}}         & {0.014}    & {}    & {0.015}       & {0.001}                  & {0.015}    & {}    & 0.024          & 0.005          & {0.022}    \\
MB-DR     & {0.014} & {0.001}                  & {0.014}    & {{}} & {{0.014}} & {0.001}                  & {{0.014}} & {}          & {{0.014}} & {0.001}                  & {0.014}    & {}    & {0.016}          & {0.001}                  & 0.016          &           & 0.027          & 0.004          & 0.026          \\
MB-Res    & {0.014} & {0.001}                  & {0.014}    & {{}} & {{0.014}} & {0.001}                  & {0.014}          & {}          & {0.015}          & {0.001}                  & {0.015}    &           & {0.016}          & {{0.001}}         & 0.016          &           & {0.023}    & {0.001} & 0.023          \\
MB-TL     & 0.055          & 0.104                        & 0.016          &                 & 0.055                & 0.104                        & 0.016                &                 & 0.055                & 0.104                        & 0.016          &           & 0.056                & 0.104                        & {0.017}    &           & 0.061          & 0.102          & {0.022}    \\
MB-PT     & 0.024          & 0.058                        & {0.013} & {}       & 0.024                & {0.058}               & {0.013}       & {}       & {0.024}       & 0.058                        & {0.013} & {} & 0.025                & 0.058                        & {0.014} & {} & 0.032          & 0.057          & {0.021}\\
\Xhline{2\arrayrulewidth}
\end{tabular}}
\end{table}

\begin{table}[ht!]
\scriptsize
\centering
\setlength{\tabcolsep}{3.5mm}
\caption{The impact of the patch size on RMSE (on Jasper Ridge), quantified as $\Delta_{\rm RMSE}={\rm RMSE}^{k\times k}-{\rm RMSE}^{3\times 3}$ ($k=\{5, 7, 9\}$) obtained using MB-DR trained from the training sets of various sizes.}\label{tab:patch_size_mbdr}
\begin{tabular}{cccccc}
\Xhline{2\arrayrulewidth}
& RMSE & & \multicolumn{3}{c}{$\Delta_{\rm RMSE}$}\\
\cline{2-2} \cline{4-6}
\multicolumn{1}{c}{\textbf{Train. size}} & \multicolumn{1}{c}{$3\times   3$} && \multicolumn{1}{c}{$5\times 5$} & \multicolumn{1}{c}{$7\times7$} & \multicolumn{1}{c}{$9\times 9$} \\
\hline
100\% & 0.014 && 0.001 & 0.001 & 0.002 \\
66\%  & 0.017 && 0.001 & 0.002 & 0.003 \\
33\%  & 0.022 && 0.003 & 0.003 & 0.007 \\
13\%  & 0.033 && 0.002 & 0.006 & 0.012 \\
6\%   & 0.042 && 0.004 & 0.015 & 0.024 \\
1\%   & 0.129 && 0.013 & 0.051 & 0.042\\
\Xhline{2\arrayrulewidth}
\end{tabular}
\end{table}

\begin{table}[ht!]
\scriptsize
\centering
\setlength{\tabcolsep}{3.5mm}
\caption{The impact of the patch size on RMSE (on Jasper Ridge), quantified as $\Delta_{\rm RMSE}={\rm RMSE}^{k\times k}-{\rm RMSE}^{3\times 3}$ ($k=\{5, 7, 9\}$) obtained using MB-Res trained from the training sets of various sizes.}
\begin{tabular}{cccccc}
\Xhline{2\arrayrulewidth}
& RMSE & & \multicolumn{3}{c}{$\Delta_{\rm RMSE}$}\\
\cline{2-2} \cline{4-6}
\multicolumn{1}{c}{\textbf{Train. size}} & \multicolumn{1}{c}{$3\times   3$} && \multicolumn{1}{c}{$5\times 5$} & \multicolumn{1}{c}{$7\times7$} & \multicolumn{1}{c}{$9\times 9$} \\
\hline
100\% & 0.015 &  & 0.240 & 0.374 & 0.359 \\
66\%  & 0.017 &  & 0.245 & 0.437 & 0.271 \\
33\%  & 0.024 &  & 0.300 & 0.403 & 0.278 \\
13\%  & 0.034 &  & 0.122 & 0.386 & 0.240 \\
6\%   & 0.059 &  & 0.232 & 0.298 & 0.302 \\
1\%   & 0.195 &  & 0.132 & 0.123 & 0.169\\
\Xhline{2\arrayrulewidth}
\end{tabular}
\end{table}

\begin{table}[ht!]
\scriptsize
\centering
\setlength{\tabcolsep}{3.5mm}
\caption{The impact of the patch size on RMSE (on Jasper Ridge), quantified as $\Delta_{\rm RMSE}={\rm RMSE}^{k\times k}-{\rm RMSE}^{3\times 3}$ ($k=\{5, 7, 9\}$) obtained using MB-TL trained from the training sets of various sizes.}
\begin{tabular}{cccccc}
\Xhline{2\arrayrulewidth}
& RMSE & & \multicolumn{3}{c}{$\Delta_{\rm RMSE}$}\\
\cline{2-2} \cline{4-6}
\multicolumn{1}{c}{\textbf{Train. size}} & \multicolumn{1}{c}{$3\times   3$} && \multicolumn{1}{c}{$5\times 5$} & \multicolumn{1}{c}{$7\times7$} & \multicolumn{1}{c}{$9\times 9$} \\
\hline
100\% & 0.151 &  & $-$0.074 & 0.020 & 0.226  \\
66\%  & 0.020 &  & 0.242  & 0.355 & 0.257  \\
33\%  & 0.275 &  & $-$0.057 & 0.067 & $-$0.011 \\
13\%  & 0.314 &  & 0.011  & 0.049 & 0.100  \\
6\%   & 0.345 &  & $-$0.012 & 0.117 & 0.082  \\
1\%   & 0.384 &  & $-$0.026 & 0.037 & 0.073 \\
\Xhline{2\arrayrulewidth}
\end{tabular}
\end{table}

\begin{table}[ht!]
\scriptsize
\centering
\setlength{\tabcolsep}{3.5mm}
\caption{The impact of the patch size on RMSE (on Jasper Ridge), quantified as $\Delta_{\rm RMSE}={\rm RMSE}^{k\times k}-{\rm RMSE}^{3\times 3}$ ($k=\{5, 7, 9\}$) obtained using MB-PT trained from the training sets of various sizes.}\label{tab:patch_size_mbpt}
\begin{tabular}{cccccc}
\Xhline{2\arrayrulewidth}
& RMSE & & \multicolumn{3}{c}{$\Delta_{\rm RMSE}$}\\
\cline{2-2} \cline{4-6}
\multicolumn{1}{c}{\textbf{Train. size}} & \multicolumn{1}{c}{$3\times   3$} && \multicolumn{1}{c}{$5\times 5$} & \multicolumn{1}{c}{$7\times7$} & \multicolumn{1}{c}{$9\times 9$} \\
\hline
100\% & 0.077 &  & 0.014  & 0.082 & 0.200  \\
66\%  & 0.016 &  & 0.001  & 0.200 & 0.156  \\
33\%  & 0.023 &  & 0.278  & 0.255 & 0.182  \\
13\%  & 0.264 &  & $-$0.129 & 0.006 & 0.009  \\
6\%   & 0.350 &  & $-$0.212 & 0.014 & 0.025  \\
1\%   & 0.412 &  & $-$0.084 & 0.033 & $-$0.026 \\
\Xhline{2\arrayrulewidth}
\end{tabular}
\end{table}

\bibliographystyle{IEEEtran}
\bibliography{ref_all}